\documentclass[sigconf]{acmart}
\AtBeginDocument{%
  \providecommand\BibTeX{{%
    \normalfont B\kern-0.5em{\scshape i\kern-0.25em b}\kern-0.8em\TeX}}}

\acmYear{}
\setcopyright{none}
\acmConference[MAD '22]{Proceedings of the 1st International Workshop on Multimedia AI against Disinformation}{June 27--30, 2022}{Newark, NJ, USA}
\acmDOI{}
\acmISBN{}

\usepackage[T1]{fontenc}
\usepackage{verbatim}
\usepackage{multirow}
\usepackage{hhline}
\usepackage[utf8x]{inputenc}
\begin{document}

\title{The MeVer DeepFake Detection Service: Lessons Learnt from Developing and Deploying in the Wild}

%\begin{comment}
\author{Spyridon Baxevanakis}
\affiliation{%
  \institution{ITI-CERTH}
  \city{Thessaloniki}
  \country{Greece}
}
\email{spirosbax@iti.gr}

\author{Giorgos Kordopatis-Zilos}
\affiliation{%
  \institution{ITI-CERTH}
  \city{Thessaloniki}
  \country{Greece}
}
\email{georgekordopatis@iti.gr}

\author{Panagiotis Galopoulos}
\affiliation{%
  \institution{ITI-CERTH}
  \city{Thessaloniki}
  \country{Greece}
}
\email{gpan@iti.gr}

\author{Lazaros Apostolidis}
\affiliation{%
  \institution{ITI-CERTH}
  \city{Thessaloniki}
  \country{Greece}
}
\email{laaposto@iti.gr}

\author{Killian Levacher}
\affiliation{%
  \institution{IBM Research}
  \city{Dublin}
  \country{Ireland}
}
\email{killian.levacher@ibm.com}

\author{Ipek B. Schlicht}
\affiliation{%
  \institution{Deutsche Welle}
  \city{Bonn/Berlin}
  \country{Germany}
}
\email{ipek.baris-schlicht@dw.com}

\author{Denis Teyssou}
\affiliation{%
  \institution{Agence France-Presse}
  \city{Paris}
  \country{France}
}
\email{denis.teyssou@afp.com}

\author{Ioannis Kompatsiaris}
\affiliation{%
  \institution{ITI-CERTH}
  \city{Thessaloniki}
  \country{Greece}
}
\email{ikom@iti.gr}

\author{Symeon Papadopoulos}
\affiliation{%
  \institution{ITI-CERTH}
  \city{Thessaloniki}
  \country{Greece}
}
\email{papadop@iti.gr}
%\end{comment}
% \author{Spiros Baxevanakis$^1$, Giorgos Kordopatis-Zilos$^1$, Panagiotis Galopoulos$^1$, Lazaros Apostolidis$^1$, Killian Levacher$^2$, Ipek Baris Schlicht$^3$, Symeon Papadopoulos$^1$}
% \affiliation{%
%   \institution{$^1$Information Technologies Institute, CERTH, Thessaloniki, Greece}
%   \country{}
% }
% \affiliation{%
%   \institution{$^2$IBM Research Europe, Dublin, Ireland}
%   \country{}
% }
% \affiliation{%
%   \institution{$^3$Deutsche Welle, Berlin, Germany}
%   \country{}
%   }
% \email{papadop@iti.gr}
% \author{Spiros Baxevanakis}
% \email{spirosbax@iti.gr}
% \orcid{0000-0001-5213-4498}

% \author{Panagiotis Galopoulos}
% % \email{spirosbax@iti.gr}
% % \orcid{0000-0001-5213-4498}

% \author{Giorgos Kordopatis-Zilos}
% % \email{spirosbax@iti.gr}
% % \orcid{0000-0001-5213-4498}

% \author{Lazaros Apostolidis}
% % \email{spirosbax@iti.gr}
% % \orcid{0000-0001-5213-4498}

% \author{Symeon Papadopoulos}
% % \email{spirosbax@iti.gr}
% % \orcid{0000-0001-5213-4498}

% \affiliation{%
%   \institution{Information Technologies Institute, CERTH}
% %   \streetaddress{P.O. Box 1212}
%   \city{Thessaloniki}
%   \country{Greece}
% %   \postcode{43017-6221}
% }

\renewcommand{\shortauthors}{The MeVer DeepFake Detection service}

\begin{abstract}
Enabled by recent improvements in generation methodologies, DeepFakes have become mainstream due to their increasingly better visual quality, the increase in easy-to-use generation tools and the rapid dissemination through social media. This fact poses a severe threat to our societies with the potential to erode social cohesion and influence our democracies. To mitigate the threat, numerous DeepFake detection schemes have been introduced in the literature but very few provide a web service that can be used in the wild. In this paper, we introduce the MeVer DeepFake detection service, a web service detecting deep learning manipulations in images and video. We present the design and implementation of the proposed
processing pipeline that involves a model ensemble scheme, and we endow the service with a model card for transparency. Experimental results show that our service performs robustly on the three benchmark datasets while being vulnerable to Adversarial Attacks. Finally, we outline our experience and lessons learned when deploying a research system into production in the hopes that it will be useful to other academic and industry teams.
\end{abstract}


% \ccsdesc[500]{Security and privacy~Authentication}
% \ccsdesc[300]{Security and privacy~Security services}

% TODO


\maketitle
\fancyhead{}

\section{Introduction}
In the fight against disinformation, facial manipulation technologies are one of the most formidable weapons that malicious actors have in their arsenal in order to deceive the public's opinion. DeepFakes stand out as perhaps the most prominent of these technologies due to the photo-realistic results and the effectiveness in social media dissemination.
% The term derives from the research field of \textit{Deep Leaning} combined with the word \textit{Fake}; thus,
A DeepFake refers to any fake image or video, typically containing facial manipulations to the displayed person(s), created using Deep Learning methods. Furthermore, non-face scenes/imagery can be the subject of DeepFakes such as satellite images \cite{zhao2021deep}.
% Enabled by advances in Computer Vision \cite{suwajanakorn_synthesizing_2017}, the first DeepFake appeared on the  Reddit platform\footnote{\url{https://www.reddit.com}} and constituted face-swapping a porn actor's face for a celebrity's.

\begin{figure}[t]
    \centering
    \includegraphics[width=0.95\linewidth]{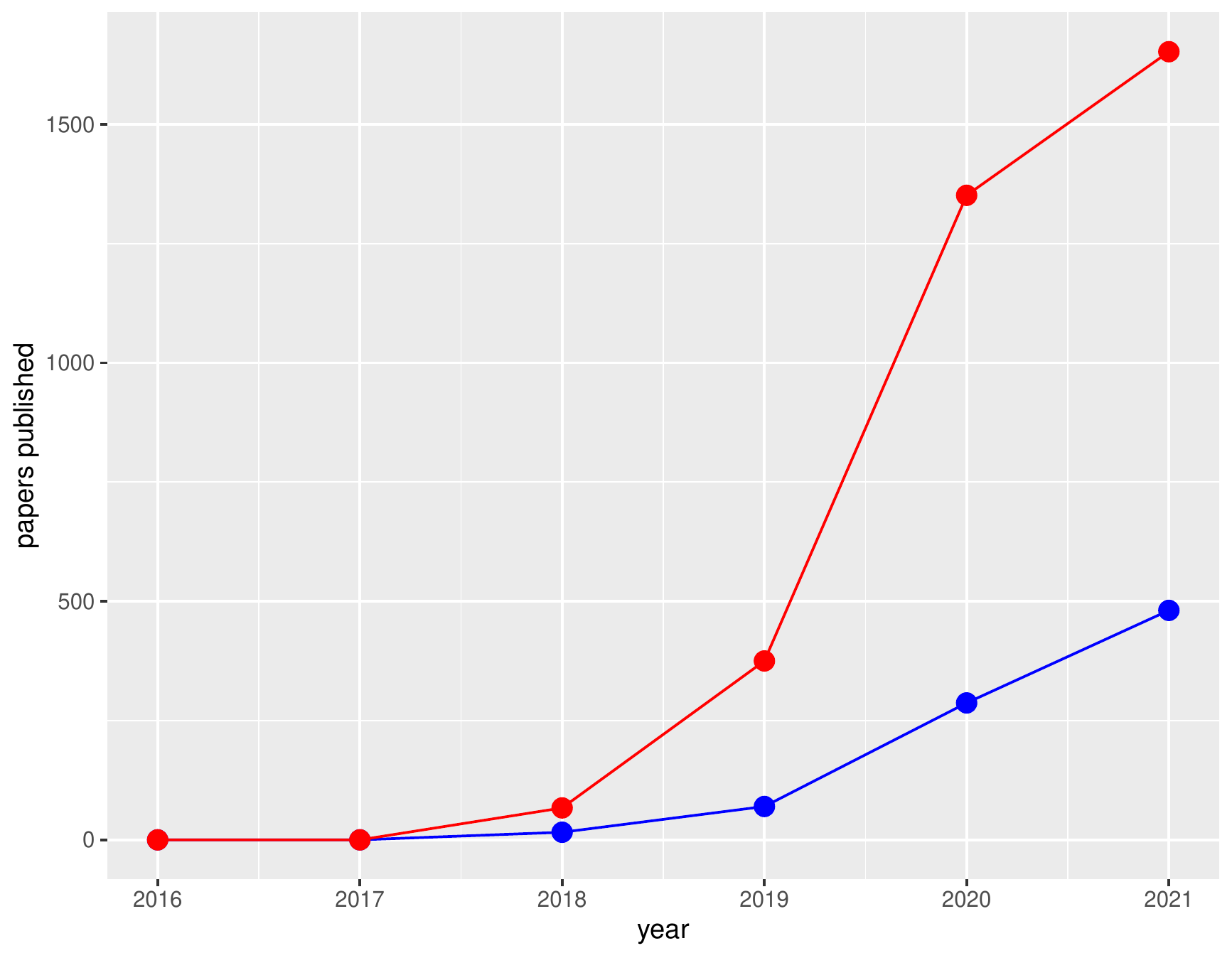}
    \caption{The red line illustrates the number of papers where the term ``DeepFake'' appears at least once in the text, while the blue line illustrates the term has to be in the title and the abstract. Data obtained from \url{https://app.dimensions.ai.}}
    \label{fig:papers}
\end{figure}

Nowadays, DeepFakes have gained popularity owing to various free and easy-to-use tools available\footnote{Examples: \url{https://faceswap.dev}, \url{https://zaodownload.com}, \url{https://facemagic.ai}} %, \url{https://hey.reface.ai}}
to anyone who wishes to create fake images and videos.
In combination with the drastic increase in quality fueled by the research in the area of image/video generation \cite{zhu_one_2021, karras_alias-free_2021, shen_interpreting_2020}, DeepFakes pose a serious threat to society with far-reaching impacts.
Some notable DeepFake examples include: a DeepFake of the US president Donald Trump in which he urges Belgian politicians to pull out of the Paris climate agreement\footnote{\url{https://www.politico.eu/article/spa-donald-trump-belgium-paris-climate-agreement-belgian-socialist-party-circulates-deep-fake-trump-video/}}, a DeepFake of Meta CEO Mark Zuckerberg in which he gives a sinister speech about the influence of Facebook on its users\footnote{\url{https://www.theguardian.com/technology/2019/jun/11/deepfake-zuckerberg-instagram-facebook}}, and a fake video of US president Barack Obama during which he insults Donald Trump\footnote{\url{https://www.theverge.com/tldr/2018/4/17/17247334/ai-fake-news-video-barack-obama-jordan-peele-buzzfeed}}.

This has attracted the interest of the multimedia community for the development of methods to tackle this threat, and as a result, the generated research in the field has skyrocketed in the last recent years.
Figure \ref{fig:papers} shows the number of papers published that mention the term \textit{DeepFake} since 2016. Furthermore, data availability has also seen such an increase in activity that in 2021 only, eight new DeepFake datasets have been released \cite{le_openforensics_2021,zhou_face_2021,kwon_kodf_2021,khodabakhsh_fake_2018, jain_improving_2021, huang_deepfake_2021, pu_deepfake_2021, frank_wavefake_2021}.

Despite these facts, the DeepFake problem remains challenging, especially in the case of novel manipulations that have not been included in the training set of DeepFake detection systems.
We argue that it is in part due to two reasons. The first relates to the challenge of training Neural Networks that are robust to out-of-distribution samples. In this context, by out-of-distribution samples we refer to DeepFakes generated with different manipulation methods than those used for training. The second reason relates to the misalignment between the synthetic datasets, developed by researchers that exhibit a strong bias towards selecting trimmed videos containing only a single face, and DeepFakes on the Internet, where videos are longer and contain many shots with multiple faces of which one or more may have been manipulated. Thus, it is evident that there is a growing need for systems that can effectively tackle these issues and mitigate the threat of DeepFakes. Such systems have to also be transparent for identifying and addressing potential issues and evaluated based on their robustness to standard adversarial attacks.

To contribute to the discussion around the problem, in this paper, we present our DeepFake detection service, its design, implementation details, and our experience deploying a multi-model system for image and video DeepFake detection in the wild. Our system receives the URL address of an image or video as input, and generates a single DeepFake probability score as output. A new input to the service triggers a multi-stage processing pipeline, including dedicated functions for the downloading and pre-processing of the input for the extraction of the contained faces. The detected faces are submitted to an ensemble scheme of five DeepFake detection models. The outputs are aggregated to derive a single probability score indicating whether the input medium contains DeepFake faces. To provide a transparent documentation for our service, we have compiled a model card. We evaluate our service on three well-known datasets and also assess the service robustness to adversarial attacks in the spirit of trustworthy AI. Finally, we document the practical challenges we faced when pivoting to a robust service API from the point of view of research code, hoping that our experience will be helpful to other academic or industry teams in the field.

\section{Related Work}

Numerous surveys and literature reviews have been published following the recent explosion in DeepFake research \cite{mirsky_creation_2021, malik_deepfake_2022, passos_review_2022, le_robust_2022}. After reviewing the creation tools and detection approaches of DeepFakes, the authors of \cite{malik_deepfake_2022} focus on the challenges for robust DeepFake detection, such as the handling of adversarial attacks.
Also, \cite{mirsky_creation_2021} reviews extensively the technical background of DeepFakes in terms of Generative Adversarial Networks (GANs), Neural Networks and Loss functions with a particular focus on Facial Reenactment techniques, such as \cite{thies_face2face_2020}.

\subsection{DeepFake Generation}
DeepFakes can be classified in five major categories based on the type of applied manipulation \cite{masood_deepfakes_2021}: (i) \textbf{FaceSwap}: This is a manipulation method where the face region of a target image is replaced with that of a source image. Most publicly available tools apply this kind of manipulation to generate DeepFakes.
(ii) \textbf{Face Reenactment (Puppet Mastery)}: In these methods, only the facial movements and expressions are transferred from a source to a target video. A seminal such method is Face2Face \cite{thies_face2face_2020}.
(iii) \textbf{Face Attribute Editing}: This manipulation modifies a selected facial attribute (e.g. eyes, skin tone, hair) while leaving the remaining face unaltered. The evolution of  Generative Adversarial Networks (GANs) in works such as \cite{he_attgan_2018} has significantly improved the realism of this kind of manipulations.
(iv) \textbf{Face Synthesis} is concerned with synthesizing entirely new images of faces and also belongs to the GAN-related family of manipulations. Notable works include StyleGAN2 \cite{karras_analyzing_2020}, used for the generation of synthetic faces in popular websites\footnote{\url{https://thispersondoesnotexist.com}}.
(v) \textbf{Lip-syncing}: In this manipulation, the mouth portion of an input video is altered to match an unrelated audio clip. Among the most influential lip-syncing works was one targeting President Barack Obama \cite{suwajanakorn_synthesizing_2017}.

For the development of our DeepFake detection service, we focus on the detection of the generated media from the first category, i.e., FaceSwap, which is the most common.

\subsection{DeepFake Detection Approaches}
Given the growing threat of tampered media to society, a lot of methods have been proposed for DeepFake detection. One of the earliest works in the field is MesoNet \cite{afchar_mesonet_2019}, where a relatively shallow Convolutional Neural Network (CNN) with five layers was proposed. In their landmark work \cite{rossler_faceforensics_2019}, the researchers benchmarked the performance of several state-of-the-art CNNs on their proposed novel FaceForensics++ dataset, showing that an XceptionNet network \cite{chollet_xception_2017} outperformed the competition.

Research since then has evolved by combining CNNs with other architectures such as Recurrent Neural Networks (RNNs) \cite{guera_deepfake_2018}, Long Short-Term Memories (LSTMs) \cite{korshunov_deepfakes_2018, mehra_deepfake_2021} or Attention heads \cite{zi_wilddeepfake_2020,zhao_learning_2020,du_towards_2020,khormali_add_2021,tran_high_2021,zhao_multi-attentional_2021}.
In \cite{bonettini_video_2020} the authors propose an ensemble of numerous CNN classifiers, based on the popular EfficientNet network \cite{tan_efficientnet_2020} in tandem with attention mechanisms and Siamese training with the goal of accurately detecting DeepFakes.
In contrast, \cite{tariq_one_2021} takes a different approach by using a video-level Convolutional LSTM-based Residual Network combined with a transfer learning training strategy to perform detection. Furthermore, the authors experimented with \textit{Merge Learning}, i.e., directly train the model with all manipulations, and \textit{Transfer Learning} or in other words using a pre-trained model from a source manipulation domain to train on a few videos from the target domain.
Following the advent of Transformer-based architectures \cite{vaswani2017attention} to the Deep Learning scene, many authors have incorporated attention mechanisms to solve the DeepFake Detection problem \cite{khan_video_2021, sun_faketransformer_2021}.
Besides these works, Capsule Networks \cite{sabour2017dynamic} have also been applied to the DeepFake problem \cite{nguyen_capsule-forensics_2019, mehra_deepfake_2021, khalil_multi-layer_2021, nguyen_capsule-forensics_2022}.
Additionally, some methods incorporate domain specific physiological signals such as head poses \cite{yang_exposing_2019} or eye blinking \cite{li_exposing_2018} in order to exploit the inconsistencies resulting from video modification.
Furthermore, since new DeepFake manipulation methods are introduced at a very rapid pace, a robust DeepFake detector should be able to generalize to examples from novel manipulation/generation models. Works that attempt to tackle the generalization problem are \cite{kim_cored_2021, kim_fretal_2021, li_face_2020}.

Following the trend in the state of the art, we build five models using the EfficientNet \cite{tan_efficientnet_2020,tan_efficientnetv2_2021} as a backbone, combined with a Transformer-based architecture, i.e., the DETR network \cite{carion_end--end_2020}.

\subsection{DeepFake Detection Services}
Seeing the threat that DeepFakes pose on society, several companies and academics have developed DeepFake detection web services.

\textit{DeepWare}\footnote{\url{https://deepware.ai/}} developed an online DeepFake scanner as well as an Android application for the identification of DeepFake videos. Their approach uses an EfficientNet-B7 \cite{tan_efficientnet_2020} pre-trained on ImageNet \cite{deng_imagenet_2009} and fine-tuned on the DFDC dataset \cite{dolhansky_deepfake_2020} that operates at frame level.
Since the dataset is imbalanced, containing approximately 20K real and 100K fake videos, they balanced it at training time by randomly selecting equal number of real or fake videos.
% They report\footnote{https://github.com/deepware/deepfake-scanner} $87.1\%$ accuracy on the DFDC dataset (TODO private test set or public test set ?) compared to $88.5\%$ which is achieved by the best performing model from the wining solution's ensemble\footnote{https://github.com/selimsef/dfdc\_deepfake\_challenge}.

\textit{DuckDuckGoose}\footnote{\url{https://duckduckgoose.ai/}} has created the DeepDetector, which is a DeepFake detection system, as well as a browser detector plugin named DeepfakeProof.
Additionally, they have created the so-called Replicant DeepFake creation system that can be used to test the reliability of biometric authentication systems. Unfortunately, they do not offer more information with regards to their model architecture, training strategy, or training data.

\textit{DeepFake-o-meter}\footnote{\url{http://zinc.cse.buffalo.edu/ubmdfl/deep-o-meter/}} is an academic non-profit work created by the University of Buffalo's Media Forensics Lab.
% in contrast with the previously mentioned services, it
Introduced in \cite{li_deepfake-o-meter_2021}, it is a web service where a user can upload a video link or file and have the DeepFake detection results be sent to the user's email. It consists of 12 DeepFake detection algorithms from the literature.

% When comparing the aforementioned services with the one presented in this work, our service is available to our trusted partners in contrast with \textit{DuckDuckGoose's} DeepFake detector product. Another key difference, is that whereas \textit{DeepFake-o-meter} implements model architectures from the literature, we developed our own model ensemble that in some cases e.g. when compared to \textit{DeepWare's} model, outperforms the competition (see Section \ref{sec:eval}).

\subsection{Trustworthy AI} \label{sec:trustai}
Autonomous AI systems are embedded into every aspect of daily life and deployed in high-impact tasks such as driving vehicles \cite{rao_deep_2018} and most currently, controlling a nuclear fusion reactor \cite{degrave_magnetic_2022}.
Thus it is evident that AI systems need to be reliable, explainable, and transparent for building trust and preventing harmful decisions. % in socio-technical systems.
In this paper, we are mostly concerned with the aspects of transparency and robustness.

Initially proposed by \cite{mitchell_model_2019}, \textit{model cards} are a form of documentation meant to accompany trained AI models. The main scope is to inform and guide end users for the proper use of the underlying tool, as well as help them interpret the output results. Among others, a model card includes details regarding the deployed model, i.e., the model's architecture or the processing pipeline that is applied given an input. It also comprises details about the data used and the process followed for the training and evaluation of the models. Additionally, model cards usually follow a versioning scheme similar to the accompanied models, where the changes from prior tool versions are described. Also, the model card facilitates the developers of such AI models so as to describe the caveats and relevant factors that may affect model performance and make recommendations for the intended use of the tool.

Adversarial attacks are a common practice that malicious actors can use to affect the performance of similar systems. These attacks come in various shapes and forms but can be categorised as: \textit{Evasion attacks} intentionally perform targeted alterations to an image or video so as to confuse a machine learning system \cite{xu_adversarial_2022} in making a wrong prediction. \textit{Poisoning attacks} \cite{aghakhani2021bullseye} attempt to alter the dataset used to train an AI model. This type of attack occurs prior to the deployment of the AI system. \textit{Extraction attacks} \cite{jagielski2020high} operate on a different dimension than previous attacks. These aim at stealing the underlying parameters of AI models and thus reproducing the same model at very little cost compared to the one invested for development. \textit{Inference attacks} \cite{choquette2021label} finally consist in identifying the characteristics of specific samples that were used to train an AI model. This can be particularly problematic when personal information was used to train a system, which could be breached and damage individuals' privacy.
A noteworthy publication is \cite{gandhi2020adversarial} where the authors evaluate the robustness of DeepFake detectors against multiple DeepFake attacks and subsequently experiment with defense methodologies against them.

To this end, in the spirit of robust and trustworthy AI, we accompany our DeepFake detection service with proper documentation, i.e., a model card, as well as evaluate it based on its robustness against adversarial attackers using evasion attacks.

\subsection{Content Authenticity Initiative}
Another interesting approach to countering the challenge of digital media manipulation is the Content Authenticity Initiative (CAI) \cite{gregory2021deepfakes}, which
%Specifically, the CAI
proposes a toolset to track the origin and manipulation history of media via an embedded Content Record.
It tracks, among other things when a specific media file was produced and by whom, what editing was performed and with what tools as well as the original file before any manipulations occurred. CAI's members include companies such as Adobe, Twitter and the New York Times.

% https://journals.sagepub.com/doi/full/10.1177/14648849211060644

\section{Service Design and Implementation}
In this section, we describe the processing pipeline and implementation of our DeepFake detection service (Section \ref{sec:pipeline}). Also, we elaborate on the deployed networks for DeepFake detection and their training process (Section \ref{sec:models}). We go into detail regarding our micro-service architecture for the service implementation (Section \ref{sec:implementation}). Finally, we present the compilation of a Model Card for the service (Section \ref{sec:model_card}).

\subsection{Processing Pipeline} \label{sec:pipeline}
Once the service receives the link of an image or video as input by the user, the following processing pipeline takes place.

\textit{Download Media}: The image/video at the URL is identified and downloaded by our custom download module that supports popular file sharing services such as Dropbox\footnote{\url{https://dropbox.com}} and Google Drive\footnote{\url{https://drive.google.com}}, as well as social media platforms like YouTube\footnote{\url{https://youtube.com}} and Twitter\footnote{\url{https://twitter.com}}.

\textit{Media Type}: If the downloaded resource is an image, then only the \textit{Face Detection} and \textit{Inference} steps that are described below are applied to get the final results. Therefore, the following steps are described below as if the resource is a video.

\textit{Video Segmentation}: During this step, a video similarity network is used in order to segment the video in multiple shots. We follow the feature extraction and similarity calculation process described in \cite{kordopatis-zilos_dns_2021}. For an input video, we extract one frame per second and derive their region-level features from a ResNet50 \cite{he2016deep} using R-MAC pooling \cite{tolias2016}. Then, we calculate the distance between consecutive frames by applying Chamfer Similarity \cite{kordopatis2019visil} on their region descriptors. Finally, we extract the peaks in the distance plot in order to determine the shot transitions. The detected shots have to be at least 1.5 seconds long. Per shot DeepFake probability scores are also displayed on the front-end, providing the user with useful information about the final video-level prediction.

\textit{Face Detection}: We apply a pre-trained MTCNN face detection network from \cite{esler_face_2022} to selected sample frames of the video (in the case of images, the face detector is applied once). We sample at most 64 unique frames per shot in order to detect and extract faces. The face detector provides squared bounding boxes that indicate the locations of the faces detected in the input image. To ensure that possible artifacts between the face and background are included, we use a margin value of 1.3, which practically means that we enlarge the detected bounding boxes by 30\% per dimension.

\textit{Face Clustering}: At this stage, the Face Clustering methodology described in \cite{charitidis_investigating_2020} is applied to all detected faces of a video shot in order to reduce the noise that is introduced by the falsely detected faces. In more detail, facial embeddings and their similarities are computed per detected face. In that way, we generate a face graph by connecting the faces with similarity greater than 0.8. We then form face clusters by extracting the graph's connected components. We filter out face clusters with only few faces, i.e., less than 20\% of the video shot's frames. % AKIS: the previous threshold doesn't make much sense. Why is the threshold relative to the shot length? if it is a very long shot then this threshold would lead to many rejections.
The remaining faces are further processed. %This step does not apply when the resource is an image.

% SPIROS: I elaborated on this step as to not give ourselves away by citing the charitidis_investigating_2020 paper and keep the present work anonymized.
% \textit{Face Clustering}:At this stage, the Face Clustering methodology described in \cite{charitidis_investigating_2020} is applied to all detected faces with the purpose of reducing the noise that is introduced by the face detector's false positives classifications. In more detail, facial embeddings are computed for every positive face detection - which is not necessarily a face since the face detector may have false positives - and then compared against each other to form similarity clusters. The hypothesis of \cite{charitidis_investigating_2020} is that false positives occur much less frequently than true positives, therefore the smaller cluster belongs to false positive detections and should be discarded. This step does not apply when the resource is an image.

\textit{Inference}: Each detected face is resized to $300 \times 300$, normalized by the ImageNet \cite{deng_imagenet_2009} mean and standard deviation, and fed to an ensemble scheme that contains five models operating in parallel. See Section \ref{sec:models} for details regarding the ensemble model. Subsequently, all five model predictions are averaged to get a DeepFake probability score per input face that ranges in $(0, 1)$.

\textit{Video-level Aggregation}:
The predictions resulting from the above processing steps are at a frame level. In order to derive an aggregated video-level DeepFake probability score, we use the following aggregation strategy:
\begin{enumerate}
    \item The face predictions of each face cluster are averaged to generate a cluster prediction.
    \item Shot predictions are derived based on the maximum prediction of their clusters.
    \item The final video-level prediction is the maximum of the shot predictions.
\end{enumerate}

\subsection{DeepFake Detection Model} \label{sec:models}

\subsubsection{Architecture} The service consists of an ensemble of the following five models with the final DeepFake probability being the ensemble's average probability.

As a backbone network for feature extraction, we used one of the EfficientNet \cite{tan_efficientnet_2020,tan_efficientnetv2_2021} networks. These are CNN models that have been automatically assembled through neural architecture search, based on a compound scaling method that uniformly scales the depth, width, and resolution of the network layers/components. We employ the EfficientNet-b4 \cite{tan_efficientnet_2020} and the EfficientNet-V2-m \cite{tan_efficientnetv2_2021}. Additionally, we use the DETR \cite{carion_end--end_2020} head on top of a backbone for some of our models. This is a Transformer Encoder-Decoder \cite{vaswani2017attention} network applied on the region-level activations generated by the backbone to aggregate them with trainable queries equal to the number of the detection classes. Since our problem is binary classification, we use only a single trainable query to derive the final prediction. Also, transformers are usually combined with positional embeddings, which can be fixed or learned. We use both for our models. Overall, we have developed the following models:
\begin{enumerate}
    \item \textbf{Model 1}: a vanilla EfficientNet-b4 \cite{tan_efficientnet_2020},
    \item \textbf{Model 2}: a Transformer head based on DETR \cite{carion_end--end_2020} with \emph{fixed} positional embeddings on top of an EfficientNet-b4 \cite{tan_efficientnet_2020},
    \item \textbf{Model 3}: a Transformer head based on DETR \cite{carion_end--end_2020} with \emph{learned} positional embeddings on top of an EfficientNet-b4 \cite{tan_efficientnet_2020},
    \item \textbf{Model 4}: a multi-head Transformer based on DETR \cite{carion_end--end_2020} on top of an EfficientNet-b4 \cite{tan_efficientnet_2020},
    \item \textbf{Model 5}: a vanilla EfficientNet-V2-m \cite{tan_efficientnetv2_2021}.
\end{enumerate}

\subsubsection{Training process}
Models 1-4 were trained on the Facebook DeepFake Detection Challenge (DFDC) dataset \cite{dolhansky_deepfake_2020} while Model 5 was trained on the WildDeepFake (WDF) dataset \cite{zi_wilddeepfake_2020}.
For the former models, we used the Adam optimizer \cite{kingma_adam_2017} with a learning rate of $10^{-4}$ and $32$ batch size for $25$ epochs respectively. The networks are initialized with pre-trained weights on ImageNet-1k \cite{deng_imagenet_2009}.
For the latter model, we trained an EfficientV2-M-in21k pre-trained on ImageNet-21k \cite{ridnik_imagenet-21k_2021} and fine-tuned using the Adam optimizer with $10^{-4}$ learning rate and $32$ batch size for $2$ epochs.
% Furthermore, both datasets were heavily augmented to avoid overfitting.
Furthermore, during training, we employ the following augmentations using the Albumentations library \cite{buslaev_albumentations_2020}: \textit{Geometric augmentations} (Rotate, HorizontalFlip), \textit{Color augmentations} (ColorJitter, ToGray), \textit{Blurring} (MotionBlur, GaussianBlur), \textit{Image Corruption} (ISONoise, CoarseDropout), \textit{External Effects} (RandomSunFlare, RandomRain). Also, for Models 1-4, we used dynamic face augmentations \cite{das2021towards}.
% \begin{itemize}
%     \item \textit{Color:} ColorJitter, ToGray
%     \item \textit{Image Corruption:} ISONoise, CoarseDropout
%     \item \textit{Blurring:} MotionBlur, GaussianBlur
%     \item \textit{Geometric:} Rotate, Horizontal Flip
%     \item \textit{External Effects:} RandomSunFlare, RandomRain.
% \end{itemize}

\subsection{Implementation} \label{sec:implementation}

\begin{figure}[t]
    \centering
    \includegraphics[width=8cm]{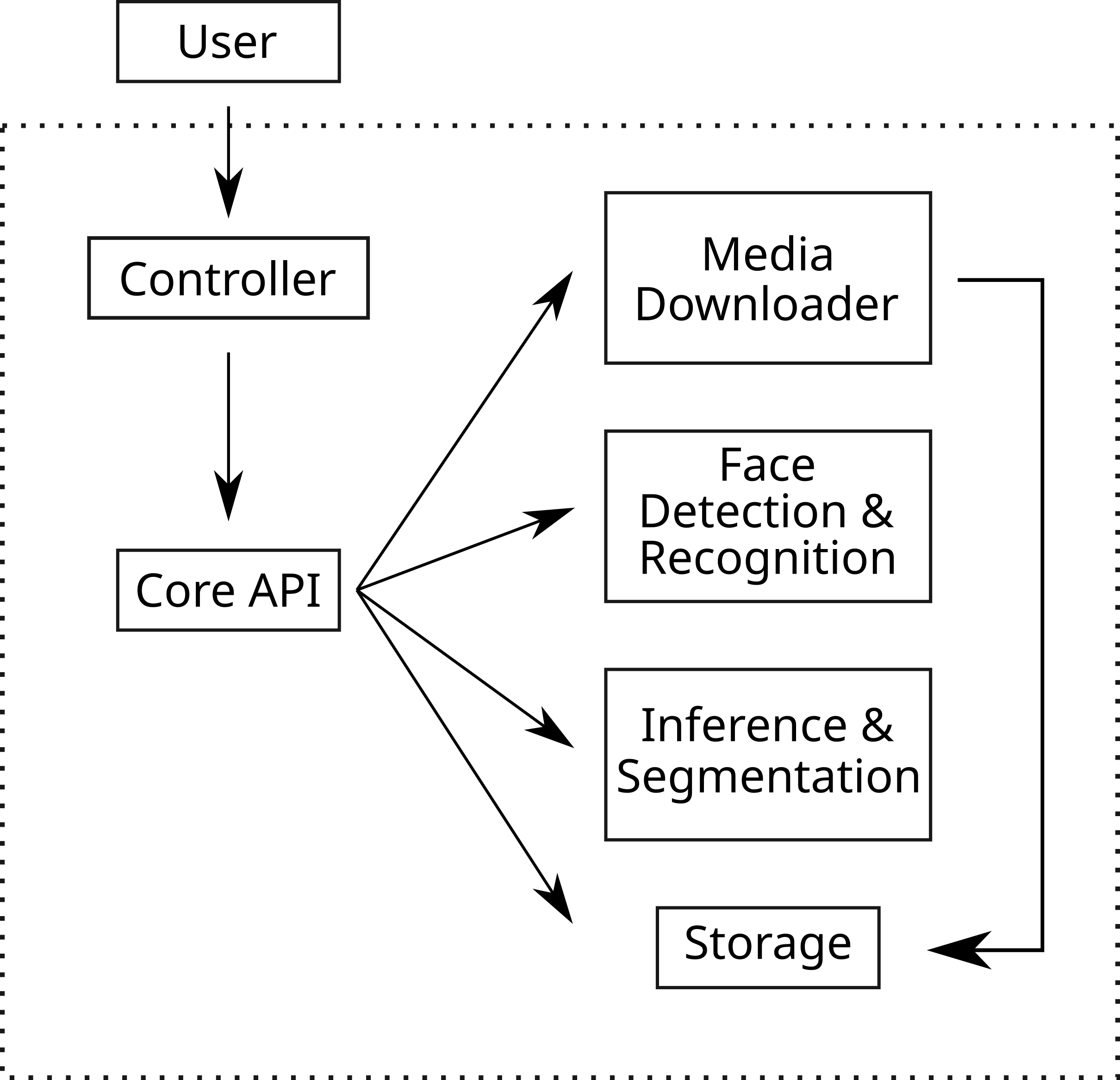}
    \caption{Service Architecture}
    \label{fig:arch}
\end{figure}

We implemented our pipeline as a set of micro-services for better modularity and  scalability. Each block in Figure \ref{fig:arch} corresponds to an independently deployed micro-service.

Our users send requests to the controller, which implements an asynchronous job API with an additional caching layer. Because the DeepFake processing can take a long time to complete, which is especially true for long videos with many segments, each request is assigned a unique job ID which is returned to the client immediately, while in the background, our system starts processing the request. The client uses the returned job ID to monitor the status of their request, and when the processing is complete, they can fetch the results. To provide low latency for repeated queries and to also reduce the strain on our system, results are stored in a Redis \cite{redis} cache.
% for a configurable amount of time.
We used FastAPI \cite{fastapi} to implement the controller's REST API, and Python-RQ \cite{python-rq} to dispatch and monitor jobs asynchronously.

Each job is implemented as a blocking HTTP call to the core service, which provides a synchronous REST API and orchestrates the necessary computations. First, it dispatches a request to the media download service that is implemented on top of Youtube-dlp \cite{youtube-dlp}. Once the video is successfully downloaded, we use OpenCV \cite{opencv} to load it and extract frames. The extracted frames are then fed to a segmentation model running on a Triton inference server \cite{triton}.

Triton is an open-source optimized inference server %, which provides an optimized back-end
for executing deep learning models on CPU or GPU. GPU memory management, batching, and model versioning are seamlessly handled. To load our models on Triton, we use the torchscript serialization of PyTorch \cite{torchscript}.

After splitting the video into shots, we use the facenet-pytorch library \cite{esler_face_2022} to run face detection and recognition on the frames of each segment. The extracted faces are then clustered based on the calculated face embeddings, and for each cluster component, we execute the DeepFake ensemble model on Triton. As each shot is independent, their processing is executed in parallel. % to speed up execution.

Finally, we aggregate the per shot and per cluster DeepFake predictions to calculate the final video-level score. We also generate ``gallery plots'', i.e. plots that present all keyframes per shot, with each keyframe drawn using a border colored based on its DeepFake score. These plots are fetched from the MinIO object store \cite{minio}, an open-source S3 compatible storage framework. %, which we use as the storage back-end for our services and specifically for storing the downloaded media and the produced plots.

% \begin{figure}[t]
%     \centering
%     \includegraphics[width=\linewidth]{figures/model_card.png}
%     \caption{Model card}
%     \label{fig:model_card}
% \end{figure}

\subsection{Model Card} \label{sec:model_card}
%As mentioned in Section \ref{sec:trustai},
We have documented our DeepFake detection service using less formal language in a Model Card format\footnote{\url{https://mever.iti.gr/deepfake/model_card.pdf}}.
% Figure \ref{fig:model_card} provides an overview of the model card compiled for our service.
%In more detail,
The model card includes a description of our service's intended use, an account of caveats and recommendations that potential novice users should take into consideration when interpreting the service results, as well as a performance evaluation over three datasets accompanied with a clear explanation of the reported metrics.
%The current version of the model card has been designed to be consumed by other researchers working on the problem of deepfake detection, as well as expert journalists and media verification companies/organizations/groups.
The compiled model card has been reviewed by AI experts from different disciplines. Based on these reviews, the current version is intended for experts having technical experience, i.e., other researchers working on the problem of DeepFake detection or media verification companies/organizations/groups. Yet, there is room for improvement for other non-technical audiences.
The compiled model card is provided in the supplementary materials.

\section{Evaluation} \label{sec:eval}
We have evaluated the performance of the presented service across three well-known DeepFake detection datasets as well as using adversarial attacks.

\subsection{Evaluation settings}

\subsubsection{Datasets}
We employ three evaluation datasets to assess the performance of our DeepFake Detection service:

\begin{itemize}
\item \textbf{FaceForensics++ (FF++)} \cite{rossler_faceforensics_2019} This is organized in two manipulation categories, \textit{Identity Swap}, implemented based on \textit{FaceSwap} and \textit{DeepFakes}, and \textit{Expression Swap}, implemented using \textit{NeuralTextures} and \textit{Face2Face}. FF++ contains 1000 real videos and 4000 fake videos derived by applying the four models on each real video.
Evaluation on FF++ provides a performance indicator on different manipulation categories and methods. Compared to more recent datasets (e.g. CelebDF, DFDC) the DeepFake quality in FF++ is visibly worse.

\item\textbf{CelebDF-V2 (CelebDF)} \cite{li_celeb-df_2020} This comprises videos from celebrity interviews that have been manipulated using improved versions of the DeepFake manipulation methods used in FF++. It consists of 590 real and 5639 fake videos.

\item\textbf{WildDeepFake (WDF)} \cite{zi_wilddeepfake_2020} In contrast to the above datasets where manipulations were generated by the dataset creators, this contains real-world DeepFakes sourced from various video-sharing websites and their corresponding real versions.
It consists of 3800 real and 3500 fake videos. Due to its real-world nature, it is considered a challenging dataset.
\end{itemize}

\subsubsection{Evaluation metrics}
Given that the evaluation datasets are imbalanced, we want to avoid skewed metrics that might favor one class or alter the datasets via sampling. Hence, we choose to report the Balanced Accuracy (BA) rather than raw Accuracy.
BA is defined as the mean of the recall computed on each class. Its possible values are in the range $0\%$-$100\%$ (higher is better).
Moreover, we report the Area Under the Curve (AUC) as it is the most often used metric in the literature. It is defined as the area under the Receiver Operating Characteristic (ROC) curve with possible values ranging from $0$ to $1$ (higher is better).

\subsubsection{Adversarial robustness}
To set up the adversarial robustness evaluation, we used IBM's Adversarial Robustness Toolbox\footnote{\url{https://adversarial-robustness-toolbox.readthedocs.io/en/latest/}} (ART).
More specifically, we used ART's \textit{PyTorchClassifier} class to wrap our model ensemble and subsequently used the Projected Gradient Descent adversarial attack \cite{madry_towards_2019} attack class. Regarding its hyperparameters, we use $\epsilon=0.2$ and $max\_iterations=5$, due to computational constraints.
To benchmark our model, for each video in each dataset, (i) we feed it to the \textit{PyTorchClassifier} unaltered, (ii) we generate an adversarial example on a frame-by-frame basis, and (iii) feed it through our classifier again.

\subsection{Experimental results}

\begin{table}[t]
\centering
\begin{tabular}{|l|c|c|c|c|}
\hline
\multirow{2}{*}{\textbf{Dataset}} & \multicolumn{2}{c|}{\textbf{MeVer}} & \multicolumn{2}{c|}{\textbf{DeepWare}} \\ \hhline{~----}
                 & \textbf{BA}     & \textbf{AUC} & \textbf{BA}     & \textbf{AUC} \\ \hline
FaceForensics++  & 70.31\% & 0.7705 & 68.77\% & 0.7681 \\ \hline
CelebDF          & 82.75\% & 0.9259 & 77.54\% & 0.9493 \\ \hline
WildDeepFake     & 84.94\% & 0.9373 & 66.96\% & 0.8646 \\ \hline
\end{tabular}
\caption[Metrics]{BA and AUC for the MeVer service (ours) and DeepWare on three datasets.}
\label{tab:datasets}
\end{table}

\subsubsection{Evaluation on different datasets}
Table \ref{tab:datasets} presents our evaluation results on the three aforementioned datasets.
Our system performs much better on the CelebDF and WDF datasets rather than the FF++, in which we observe an average $13\%$ performance drop in terms of BA.
Specifically, on the WDF dataset, we achieve an $84.94\%$ BA which is close to other state-of-the-art methods \cite{zi_wilddeepfake_2020} and is expected since our ensemble's fifth model was trained on this dataset (see Section \ref{sec:models}). In the case of the CelebDF dataset, even though none of our models have been trained with this dataset, we achieve an $82.75\%$ BA. Additionally, we compare our system with the publicly available model by DeepWare\footnote{\url{https://github.com/deepware/deepfake-scanner}}. We follow the same settings as used for our models for pre-processing, which are slightly different from those used by the original authors. The two systems perform comparably in FF++ and CelebDF, with our service having a small but clear edge. Our system significantly outperforms its competitor in WDF, which is expected since it has been used for training.

\subsubsection{Evaluation on different manipulations}
To delve into the performance discrepancy between FF++ and the other two datasets, we performed a more extensive evaluation on the FF++ dataset in terms of manipulation type; this is presented in Table \ref{tab:ff}.
Our models are considerably better at detecting \textit{FaceSwap} and \textit{DeepFake} manipulations than \textit{NeuralTextures} and \textit{Face2Face} manipulations.
The former two belong to the \textit{Identity Swap} manipulation category while the latter two are examples of \textit{Expression Swapping} \cite{masood_deepfakes_2021}.
It can be argued that this is due to our training data lacking \textit{Expression Swapping} examples; therefore, we expect our service to perform better on \textit{Identity Swap} manipulations.

%The performance drop off in FF++, is likely due to our training data lacking \textit{Expression Swap} examples, which is one of the two manipulation categories of that dataset. In Table \ref{tab:ff}, we observe worse performance in \textit{NeuralTextures} and \textit{Face2Face} manipulations (both of type \textit{Expression Swap}), hence we recommend the current version of the service to be used for the detection of Identify Swap rather than Expression Swap.

\begin{table}[t]
\centering
\begin{tabular}{|l|c|c|}
\hline
\textbf{Manipulation} & \textbf{BA} & \textbf{AUC} \\ \hline
FaceSwap       & 78.40\% & 0.8674            \\ \hline
DeepFakes      & 86.20\% & 0.9468            \\ \hline
NeuralTextures & 57.65\% & 0.6276            \\ \hline
Face2Face      & 59.02\% & 0.6402            \\ \hline
\end{tabular}
\caption[Metrics]{BA and AUC for each manipulation in FF++.}
\label{tab:ff}
\end{table}

\subsubsection{Adversarial robustness} \label{sec:adv}
Table \ref{tab:adv} illustrates the service performance in terms of its robustness to adversarial attacks with the PGD attack with three different output normalization settings.
Even though PGD is a white-box attack, meaning that the attacker would need access to the weights of all the ensemble models, we maintain that in the spirit of Reliable AI, it is preferable to consider such a worst-case scenario.
All attacks try to fool the detector into assessing that the input media are real.
The norm-1 attack does not have any noticeable effect on the performance in comparison to the original performance from Table \ref{tab:datasets}.
However, the norm-2 attack considerably affects the detection accuracy, even though the models still retain decent performance.
The strongest norm-inf attack highlights the susceptibility of our model to such attacks as it can no longer distinguish between real and DeepFake videos.
Yet, traces of an adversarial attack are visible with the naked eye in images attacked with the norm-inf.

\begin{table} [t]
\centering
\begin{tabular}{|l|c|c|c|c|}
\hline
\textbf{Dataset}        & \textbf{norm-1} & \textbf{norm-2} & \textbf{norm-inf} \\ \hline
FaceForensics++         & 70.31\%               & 64.04\%               & 50.53\%                 \\ \hline
CelebDF                 & 82.75\%               & 76.01\%               & 50.00\%                 \\ \hline
WildDeepFake            & 84.94\%               & 63.04\%               & 50.00\%                 \\ \hline
\end{tabular}
\caption[Adversarial Metrics]{BA on three datasets attacked with the PGD adversarial attack with three output normalization setting.}
\label{tab:adv}
\end{table}

\section{Deploying in the Wild}
Transferring our code from research to production proved to be challenging for multiple reasons. We go through the challenges we faced hoping that our experience will be of use to other academics deploying their research to more real-world settings (Section \ref{sec:practical}). Also, we have built a User Interface for demonstration purposes (Section \ref{sec:user_interface}). We finally discuss our versioning process (Section \ref{sec:versioning}) and some considerations regarding access and availability of the tool (Section \ref{sec:availability})

\subsection{Practical Challenges} \label{sec:practical}
During our research, we paid little attention to error handling, and as expected, that was not enough to implement a robust API with helpful explanations when things go wrong. Our micro-service design, however, helped enforce modularity with clear error boundaries. For example, Triton was responsible for GPU-related issues, separate micro-services were dedicated to downloading, face detection and recognition pipelines, while MinIO \cite{minio} offered a storage abstraction. Internally in our APIs, we consistently checked for expected errors and made use of the standard application/problem + JSON content type\footnote{\url{https://datatracker.ietf.org/doc/rfc7807}} to propagate them externally when necessary.

\begin{figure}[t]
    \centering
    \includegraphics[width=6cm]{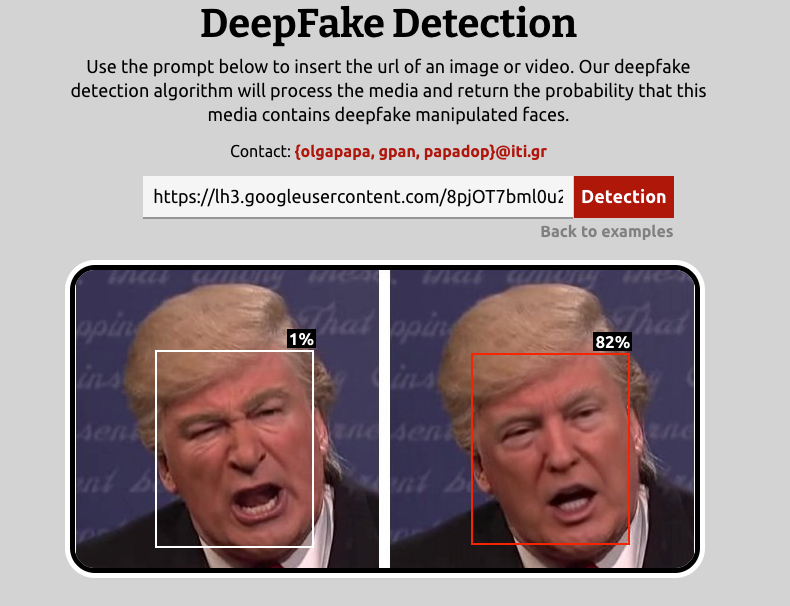}
    \caption{Image Analysis User Interface for the DeepFake detection service}
    \label{fig:image_ui}
\end{figure}

We also faced a number of difficulties with the video downloading process. %To provide a smooth user experience,
We tried to support as many video sources as we could, the most popular being YouTube, Twitter, and Facebook. Initially, our download back-end depended on Youtube-dl \cite{youtube-dl}; however, we experienced very slow download speeds. We averaged around 50KB/s, which for a 10MB video would translate to about three and a half minutes of download time, adding significantly to the total latency. Favoring high-definition video versions - since video quality is one of the most important factors for getting higher accuracy results - would further exacerbate the issue. %also did not make things easier.
Fortunately, Youtube-dlp \cite{youtube-dlp}, a fork of Youtube-dl, allowed us to consistently achieve much higher download speeds, averaging around 700KB/s. Another complication was that downloading using our public IP could result in additional throttling or even denial of service. For this reason, we chose to run our downloader behind a TOR\footnote{\url{https://torproject.org}} proxy to provide us with anonymity. Finally, an issue that we faced and still have not resolved is that Facebook downloads are rarely successful without Facebook user authentication. %using credentials.

\subsection{User Interface} \label{sec:user_interface}

\begin{figure}[t]
    \centering
    \includegraphics[width=4cm]{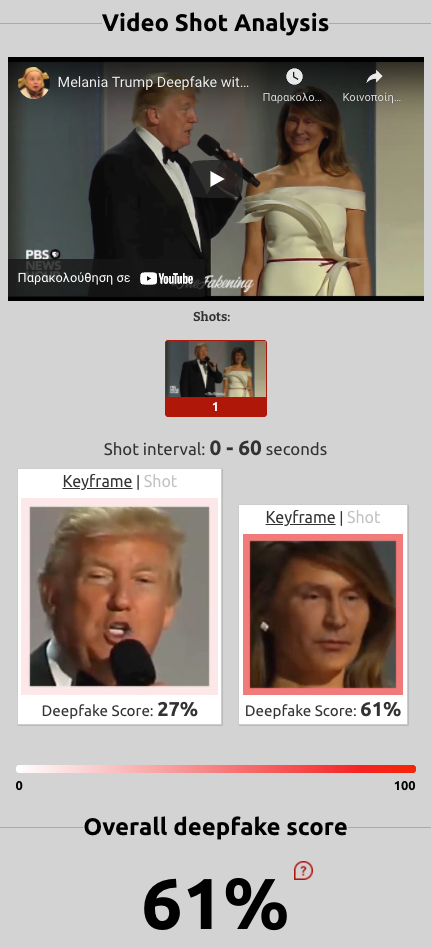}
    \caption{Video Analysis User Interface for the DeepFake detection service}
    \label{fig:video_ui}
\end{figure}

\begin{figure}[t]
    \centering
    \includegraphics[width=9cm]{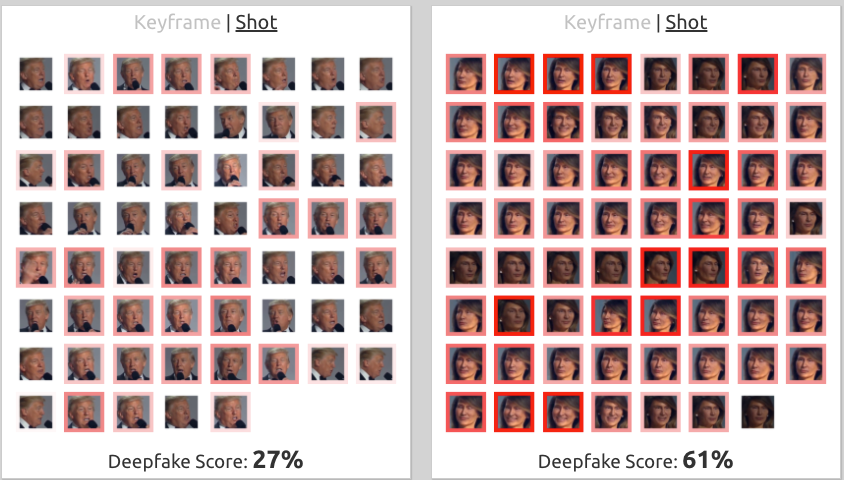}
    \caption{Frame collage view from the selected shot during Video analysis}
    \label{fig:collage_ui}
\end{figure}

\begin{figure}[t]
    \centering
    \includegraphics[width=6.2cm]{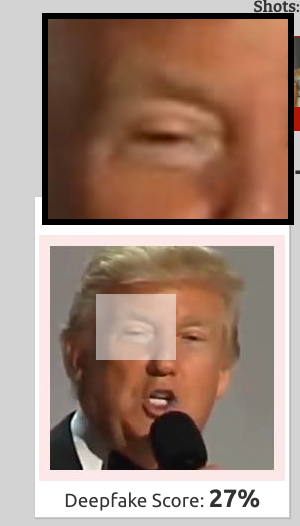}
    \caption{Hover-to-Zoom functionality in Image and Video analysis}
    \label{fig:zoom_ui}
\end{figure}

In terms of User Interface (UI), our service provides two modes: Image and Video analysis.
In Image analysis, a DeepFake probability score is presented for each detected face and displayed on top of the face's bounding box as shown in Figure \ref{fig:image_ui}. An example of video analysis can be seen in Figure \ref{fig:video_ui}. First, an embedded player allows users to playback the video, and second, the UI displays a shot selector as well as a shot interval that the user may choose to inspect the DeepFake probabilities for a specific shot.
By default, the initially selected shot is the one with the highest DeepFake probability.
Once a shot is selected, the UI displays a window for each detected face accompanied with its corresponding DeepFake probability.
The window provides the ``Keyframe'' (Figure \ref{fig:collage_ui}) and ``Shot'' (default) views which show a selected frame and a collage of frames from the shot, respectively.
Furthermore, the window allows the user to use the ``Hover-to-Zoom'' functionality for closer inspection of each view, as seen in Figure \ref{fig:zoom_ui}.
Last, the ``Overall'' DeepFake score is displayed at the bottom of the page, which results from applying the aggregation strategy described in Section \ref{sec:pipeline}.

\subsection{Versioning} \label{sec:versioning}
As with every software project, it is important to have a clear versioning scheme.
For the presented service, we have decided to use an adaptation of the Semantic Versioning 2 scheme\footnote{\url{https://semver.org/}}. In particular, we follow the $x$.$y$.$z$ scheme where:
\begin{itemize}
    \item $x$: is used for backward-incompatible changes, i.e., change of the output of the service.
    \item $y$: refers to changes in the processing pipeline such as the video segmentation methodology, the deployed models used, and others as described in Section \ref{sec:pipeline},
    \item $z$: is reserved for minor changes such as the aggregation strategy or changes in the model's input dimensions and minor bug fixes.
\end{itemize}

\subsection{Availability} \label{sec:availability}
% + similar models
We have chosen to keep the code and the model's weights private and to require user credentials for granting access to our UI.
This is due to two main reasons.
First, we argue that since our models are vulnerable to adversarial attacks - as shown in Section \ref{sec:adv} - it is essential to protect the service from white-box attacks that would be very easy if the internals were public. We believe that this action will deter the large majority of malicious actors from performing white-box adversarial attacks.
% Hence, we have sought to maximize the effort needed to generate a
Second, given the scarcity of publicly accessible DeepFake detection services, we lack the computational resources to handle the potentially high traffic from end users, and other complications including Denial of Service (DoS) attacks.
%Also our ... is ... to avoid bad actors.
For the above reasons, %the service and UI are not publicly available, and
we grant access to the service and UI only to trusted partners upon request. Finally, our service has been integrated and is accessible through the InViD-WeVerify Verification plugin \cite{teyssou2017invid} (for approved users) and the Truly Media\footnote{https://www.truly.media/} application.

\section{Conclusion}
In this work, we introduced the MeVer DeepFake detection service, a complex multi-model system that detects DeepFake videos and images.
We discussed the overall processing pipeline, including a number of pre-processing steps.
Also, we presented the model architectures and training processes for the deployed models as well as implementation details for the service.
The service has been evaluated on three well-known datasets: FaceForensics++, Celeb-DF, and WildDeepFake. For Celeb-DF and WildDeepFake, our service performed robustly and better compared to the publicly available DeepWare model. For FF++, evaluation by manipulation type revealed that our service performed robustly only in Identity Swap manipulations. % in two out of three datasets, but it does not detect all manipulation categories with the same accuracy.
In the spirit of \textit{Trustworthy AI}, we also performed an Adversarial Robustness evaluation, and we provided a model card for the service.
From the results of the adversarial evaluation, we observe a vulnerability to the \textit{Projected Gradient Descent} attack, which opens new directions for future research.
Last but not least, we discuss at length the practical challenges we faced moving from a research codebase to a real-world system that would need to be robust to arbitrary media content from the Internet. % on the web.

% GIORGOS: this might be redundant
% AKIS: Agree!
%Multiple DeepFake detection schemes have been proposed in the literature, with perhaps most promising being hybrid approaches that combine CNNs with attention mechanisms \cite{khormali_add_2021, zhao_multi-attentional_2021}, Capsule Networks \cite{mehra_deepfake_2021, nguyen_capsule-forensics_2022} or Transformers \cite{khan_video_2021, sun_faketransformer_2021}.
%One problem that has not received much attention as of yet is making the DeepFake detections robust to Adversarial Attacks. Using an Adversarial attack, an attacker can easily fool a DeepFake detector into predicting that a DeepFake video is real, thus nullifying its capacity as a verification tool. This remains a major shortcoming of DeepFake detection systems.
%Apart from this, when taking into account the recent explosion in DeepFake generation research interest, it is evident that the generalization ability of DeepFake detectors is paramount to deterring the spread of manipulated media.
%Thus we believe that future literature should focus on cross-dataset and robustness evaluations.

%With regards to the service presented,
In the future, we plan to improve the detection accuracy by continuously employing the most recent advancements in the field, i.e., by using better datasets for training and evaluation and using state-of-the-art model architectures. In addition, we plan to experiment with various promising Deep Learning architectures and training techniques in order to keep up with the ever-increasing visual quality of DeepFakes. Also, we plan to enhance our service with methods for the detection of fully synthetically generated faces (e.g. based on StyleGAN3 \cite{karras_alias-free_2021}), which is not supported in the current version. Furthermore, we plan to compile more versions of model cards targeted at wider non-technical audiences, e.g., journalists or business managers.
Last but not least, we are committed to maintaining our service as well as improving on the underlying API and User Interface.

\section*{Acknowledgements} This work has been supported by the AI4Media H2020 project, partially funded by the European Commission under contract number 951911, and the US Paris Tech Challenge award, which was funded by the US Department of State Global Engagement Center under contract number SGECPD18CA0024. Also, we would like to thank Birgit Gray from DW and Beat Buesser from IBM for their valuable feedback and support.

\balance
\bibliographystyle{ACM-Reference-Format}
\bibliography{refs}


\begin{thebibliography}{84}

%%% the Web address.

\ifx \showCODEN    \undefined \def \showCODEN     #1{\unskip}     \fi
\ifx \showDOI      \undefined \def \showDOI       #1{#1}\fi
\ifx \showISBNx    \undefined \def \showISBNx     #1{\unskip}     \fi
\ifx \showISBNxiii \undefined \def \showISBNxiii  #1{\unskip}     \fi
\ifx \showISSN     \undefined \def \showISSN      #1{\unskip}     \fi
\ifx \showLCCN     \undefined \def \showLCCN      #1{\unskip}     \fi
\ifx \shownote     \undefined \def \shownote      #1{#1}          \fi
\ifx \showarticletitle \undefined \def \showarticletitle #1{#1}   \fi
\ifx \showURL      \undefined \def \showURL       {\relax}        \fi
% invisible to TeX
\providecommand\bibfield[2]{#2}
\providecommand\bibinfo[2]{#2}
\providecommand\natexlab[1]{#1}
\providecommand\showeprint[2][]{arXiv:#2}

\bibitem[Afchar et~al\mbox{.}(2019)]%
        {afchar_mesonet_2019}
\bibfield{author}{\bibinfo{person}{Darius Afchar}, \bibinfo{person}{Vincent
  Nozick}, \bibinfo{person}{Junichi Yamagishi}, {and} \bibinfo{person}{Isao
  Echizen}.} \bibinfo{year}{2019}\natexlab{}.
\newblock \showarticletitle{{MesoNet}: {A} compact facial video forgery
  detection network}.
\newblock \bibinfo{journal}{\emph{10th IEEE International Workshop on
  Information Forensics and Security, WIFS 2018}} (\bibinfo{year}{2019}).
\newblock
\showISSN{2331-8422}
\urldef\tempurl%
\url{https://doi.org/10.1109/WIFS.2018.8630761}
\showDOI{\tempurl}
\newblock
\shownote{ISBN: 9781538665367}.


\bibitem[Aghakhani et~al\mbox{.}(2021)]%
        {aghakhani2021bullseye}
\bibfield{author}{\bibinfo{person}{Hojjat Aghakhani}, \bibinfo{person}{Dongyu
  Meng}, \bibinfo{person}{Yu-Xiang Wang}, \bibinfo{person}{Christopher
  Kruegel}, {and} \bibinfo{person}{Giovanni Vigna}.}
  \bibinfo{year}{2021}\natexlab{}.
\newblock \showarticletitle{Bullseye polytope: A scalable clean-label poisoning
  attack with improved transferability}. In \bibinfo{booktitle}{\emph{2021 IEEE
  European Symposium on Security and Privacy (EuroS\&P)}}. IEEE,
  \bibinfo{pages}{159--178}.
\newblock


\bibitem[Bonettini et~al\mbox{.}(2020)]%
        {bonettini_video_2020}
\bibfield{author}{\bibinfo{person}{Nicolò Bonettini}, \bibinfo{person}{Luca
  Bondi}, \bibinfo{person}{Edoardo~Daniele Cannas}, \bibinfo{person}{Paolo
  Bestagini}, \bibinfo{person}{Sara Mandelli}, {and} \bibinfo{person}{Stefano
  Tubaro}.} \bibinfo{year}{2020}\natexlab{}.
\newblock \showarticletitle{Video face manipulation detection through ensemble
  of {CNNs}}.
\newblock \bibinfo{journal}{\emph{Proceedings - International Conference on
  Pattern Recognition}} (\bibinfo{year}{2020}), \bibinfo{pages}{5012--5019}.
\newblock
\showISSN{10514651}
\urldef\tempurl%
\url{https://doi.org/10.1109/ICPR48806.2021.9412711}
\showDOI{\tempurl}
\newblock
\shownote{ISBN: 9781728188089}.


\bibitem[Buslaev et~al\mbox{.}(2020)]%
        {buslaev_albumentations_2020}
\bibfield{author}{\bibinfo{person}{Alexander Buslaev},
  \bibinfo{person}{Vladimir~I. Iglovikov}, \bibinfo{person}{Eugene
  Khvedchenya}, \bibinfo{person}{Alex Parinov}, \bibinfo{person}{Mikhail
  Druzhinin}, {and} \bibinfo{person}{Alexandr~A. Kalinin}.}
  \bibinfo{year}{2020}\natexlab{}.
\newblock \showarticletitle{Albumentations: {Fast} and {Flexible} {Image}
  {Augmentations}}.
\newblock \bibinfo{journal}{\emph{Information}} \bibinfo{volume}{11},
  \bibinfo{number}{2} (\bibinfo{date}{Feb.} \bibinfo{year}{2020}),
  \bibinfo{pages}{125}.
\newblock
\showISSN{2078-2489}
\urldef\tempurl%
\url{https://doi.org/10.3390/info11020125}
\showDOI{\tempurl}
\newblock
\shownote{Number: 2 Publisher: Multidisciplinary Digital Publishing Institute}.


\bibitem[Carion et~al\mbox{.}(2020)]%
        {carion_end--end_2020}
\bibfield{author}{\bibinfo{person}{Nicolas Carion}, \bibinfo{person}{Francisco
  Massa}, \bibinfo{person}{Gabriel Synnaeve}, \bibinfo{person}{Nicolas
  Usunier}, \bibinfo{person}{Alexander Kirillov}, {and} \bibinfo{person}{Sergey
  Zagoruyko}.} \bibinfo{year}{2020}\natexlab{}.
\newblock \showarticletitle{End-to-end object detection with transformers}. In
  \bibinfo{booktitle}{\emph{European conference on computer vision}}. Springer,
  \bibinfo{pages}{213--229}.
\newblock


\bibitem[Charitidis et~al\mbox{.}(2020)]%
        {charitidis_investigating_2020}
\bibfield{author}{\bibinfo{person}{Polychronis Charitidis},
  \bibinfo{person}{Giorgos Kordopatis-Zilos}, \bibinfo{person}{Symeon
  Papadopoulos}, {and} \bibinfo{person}{Ioannis Kompatsiaris}.}
  \bibinfo{year}{2020}\natexlab{}.
\newblock \showarticletitle{Investigating the impact of pre-processing and
  prediction aggregation on the deepfake detection task}.
\newblock \bibinfo{journal}{\emph{Proceedings of the 2020 Truth and Trust
  Online}} (\bibinfo{year}{2020}), \bibinfo{pages}{44--54}.
\newblock


\bibitem[Chollet(2017)]%
        {chollet_xception_2017}
\bibfield{author}{\bibinfo{person}{François Chollet}.}
  \bibinfo{year}{2017}\natexlab{}.
\newblock \showarticletitle{Xception: {Deep} learning with depthwise separable
  convolutions}.
\newblock \bibinfo{journal}{\emph{Proceedings - 30th IEEE Conference on
  Computer Vision and Pattern Recognition, CVPR 2017}}
  \bibinfo{volume}{2017-Janua} (\bibinfo{year}{2017}),
  \bibinfo{pages}{1800--1807}.
\newblock
\urldef\tempurl%
\url{https://doi.org/10.1109/CVPR.2017.195}
\showDOI{\tempurl}
\newblock
\shownote{ISBN: 9781538604571}.


\bibitem[Choquette-Choo et~al\mbox{.}(2021)]%
        {choquette2021label}
\bibfield{author}{\bibinfo{person}{Christopher~A Choquette-Choo},
  \bibinfo{person}{Florian Tramer}, \bibinfo{person}{Nicholas Carlini}, {and}
  \bibinfo{person}{Nicolas Papernot}.} \bibinfo{year}{2021}\natexlab{}.
\newblock \showarticletitle{Label-only membership inference attacks}. In
  \bibinfo{booktitle}{\emph{International Conference on Machine Learning}}.
  PMLR, \bibinfo{pages}{1964--1974}.
\newblock


\bibitem[Das et~al\mbox{.}(2021)]%
        {das2021towards}
\bibfield{author}{\bibinfo{person}{Sowmen Das}, \bibinfo{person}{Selim
  Seferbekov}, \bibinfo{person}{Arup Datta}, \bibinfo{person}{Md Islam},
  \bibinfo{person}{Md Amin}, {et~al\mbox{.}}} \bibinfo{year}{2021}\natexlab{}.
\newblock \showarticletitle{Towards solving the deepfake problem: An analysis
  on improving deepfake detection using dynamic face augmentation}. In
  \bibinfo{booktitle}{\emph{Proceedings of the IEEE/CVF International
  Conference on Computer Vision}}. \bibinfo{pages}{3776--3785}.
\newblock


\bibitem[Degrave et~al\mbox{.}(2022)]%
        {degrave_magnetic_2022}
\bibfield{author}{\bibinfo{person}{Jonas Degrave}, \bibinfo{person}{Federico
  Felici}, \bibinfo{person}{Jonas Buchli}, \bibinfo{person}{Michael Neunert},
  \bibinfo{person}{Brendan Tracey}, \bibinfo{person}{Francesco Carpanese},
  \bibinfo{person}{Timo Ewalds}, \bibinfo{person}{Roland Hafner},
  \bibinfo{person}{Abbas Abdolmaleki}, \bibinfo{person}{Diego de~las Casas},
  \bibinfo{person}{Craig Donner}, \bibinfo{person}{Leslie Fritz},
  \bibinfo{person}{Cristian Galperti}, \bibinfo{person}{Andrea Huber},
  \bibinfo{person}{James Keeling}, \bibinfo{person}{Maria Tsimpoukelli},
  \bibinfo{person}{Jackie Kay}, \bibinfo{person}{Antoine Merle},
  \bibinfo{person}{Jean-Marc Moret}, \bibinfo{person}{Seb Noury},
  \bibinfo{person}{Federico Pesamosca}, \bibinfo{person}{David Pfau},
  \bibinfo{person}{Olivier Sauter}, \bibinfo{person}{Cristian Sommariva},
  \bibinfo{person}{Stefano Coda}, \bibinfo{person}{Basil Duval},
  \bibinfo{person}{Ambrogio Fasoli}, \bibinfo{person}{Pushmeet Kohli},
  \bibinfo{person}{Koray Kavukcuoglu}, \bibinfo{person}{Demis Hassabis}, {and}
  \bibinfo{person}{Martin Riedmiller}.} \bibinfo{year}{2022}\natexlab{}.
\newblock \showarticletitle{Magnetic control of tokamak plasmas through deep
  reinforcement learning}.
\newblock \bibinfo{journal}{\emph{Nature}} \bibinfo{volume}{602},
  \bibinfo{number}{7897} (\bibinfo{date}{Feb.} \bibinfo{year}{2022}),
  \bibinfo{pages}{414--419}.
\newblock
\showISSN{1476-4687}
\urldef\tempurl%
\url{https://doi.org/10.1038/s41586-021-04301-9}
\showDOI{\tempurl}
\newblock
\shownote{Number: 7897 Publisher: Nature Publishing Group}.


\bibitem[Deng et~al\mbox{.}(2009)]%
        {deng_imagenet_2009}
\bibfield{author}{\bibinfo{person}{Jia Deng}, \bibinfo{person}{Wei Dong},
  \bibinfo{person}{Richard Socher}, \bibinfo{person}{Li-Jia Li},
  \bibinfo{person}{Kai Li}, {and} \bibinfo{person}{Li Fei-Fei}.}
  \bibinfo{year}{2009}\natexlab{}.
\newblock \showarticletitle{{ImageNet}: {A} large-scale hierarchical image
  database}. In \bibinfo{booktitle}{\emph{2009 {IEEE} {Conference} on
  {Computer} {Vision} and {Pattern} {Recognition}}}. \bibinfo{pages}{248--255}.
\newblock
\urldef\tempurl%
\url{https://doi.org/10.1109/CVPR.2009.5206848}
\showDOI{\tempurl}
\newblock
\shownote{ISSN: 1063-6919}.


\bibitem[Dolhansky et~al\mbox{.}(2020)]%
        {dolhansky_deepfake_2020}
\bibfield{author}{\bibinfo{person}{Brian Dolhansky}, \bibinfo{person}{Joanna
  Bitton}, \bibinfo{person}{Ben Pflaum}, \bibinfo{person}{Jikuo Lu},
  \bibinfo{person}{Russ Howes}, \bibinfo{person}{Menglin Wang}, {and}
  \bibinfo{person}{Cristian~Canton Ferrer}.} \bibinfo{year}{2020}\natexlab{}.
\newblock \showarticletitle{The {DeepFake} {Detection} {Challenge} ({DFDC})
  {Dataset}}.
\newblock  (\bibinfo{year}{2020}).
\newblock
\newblock
\shownote{arXiv: 2006.07397}.


\bibitem[Driessen(2021)]%
        {python-rq}
\bibfield{author}{\bibinfo{person}{Vincent Driessen}.}
  \bibinfo{year}{2021}\natexlab{}.
\newblock \bibinfo{title}{Python-RQ}.
\newblock \bibinfo{howpublished}{\url{https://python-rq.org/docs/}}.
\newblock


\bibitem[Du et~al\mbox{.}(2020)]%
        {du_towards_2020}
\bibfield{author}{\bibinfo{person}{Mengnan Du}, \bibinfo{person}{Shiva
  Pentyala}, \bibinfo{person}{Yuening Li}, {and} \bibinfo{person}{Xia Hu}.}
  \bibinfo{year}{2020}\natexlab{}.
\newblock \showarticletitle{Towards generalizable deepfake detection with
  locality-aware autoencoder}. In \bibinfo{booktitle}{\emph{Proceedings of the
  29th ACM International Conference on Information \& Knowledge Management}}.
  \bibinfo{pages}{325--334}.
\newblock


\bibitem[Esler(2022)]%
        {esler_face_2022}
\bibfield{author}{\bibinfo{person}{Tim Esler}.}
  \bibinfo{year}{2022}\natexlab{}.
\newblock \bibinfo{title}{Face {Recognition} {Using} {Pytorch}}.
\newblock
\newblock
\urldef\tempurl%
\url{https://github.com/timesler/facenet-pytorch}
\showURL{%
\tempurl}
\newblock
\shownote{original-date: 2019-05-25T01:29:24Z}.


\bibitem[Frank and Schönherr(2021)]%
        {frank_wavefake_2021}
\bibfield{author}{\bibinfo{person}{Joel Frank} {and} \bibinfo{person}{Lea
  Schönherr}.} \bibinfo{year}{2021}\natexlab{}.
\newblock \bibinfo{title}{{WaveFake}: {A} data set to facilitate audio
  {DeepFake} detection}.
\newblock
\newblock
\urldef\tempurl%
\url{https://doi.org/10.5281/zenodo.5642694}
\showDOI{\tempurl}
\newblock
\shownote{Type: dataset}.


\bibitem[Gandhi and Jain(2020)]%
        {gandhi2020adversarial}
\bibfield{author}{\bibinfo{person}{Apurva Gandhi} {and} \bibinfo{person}{Shomik
  Jain}.} \bibinfo{year}{2020}\natexlab{}.
\newblock \showarticletitle{Adversarial perturbations fool deepfake detectors}.
  In \bibinfo{booktitle}{\emph{2020 international joint conference on neural
  networks (IJCNN)}}. IEEE, \bibinfo{pages}{1--8}.
\newblock


\bibitem[Gregory(2021)]%
        {gregory2021deepfakes}
\bibfield{author}{\bibinfo{person}{Sam Gregory}.}
  \bibinfo{year}{2021}\natexlab{}.
\newblock \showarticletitle{Deepfakes, misinformation and disinformation and
  authenticity infrastructure responses: Impacts on frontline witnessing,
  distant witnessing, and civic journalism}.
\newblock \bibinfo{journal}{\emph{Journalism}} (\bibinfo{year}{2021}),
  \bibinfo{pages}{14648849211060644}.
\newblock


\bibitem[Güera and Delp(2018)]%
        {guera_deepfake_2018}
\bibfield{author}{\bibinfo{person}{David Güera} {and}
  \bibinfo{person}{Edward~J. Delp}.} \bibinfo{year}{2018}\natexlab{}.
\newblock \showarticletitle{Deepfake {Video} {Detection} {Using} {Recurrent}
  {Neural} {Networks}}. In \bibinfo{booktitle}{\emph{2018 15th {IEEE}
  {International} {Conference} on {Advanced} {Video} and {Signal} {Based}
  {Surveillance} ({AVSS})}}. \bibinfo{pages}{1--6}.
\newblock
\urldef\tempurl%
\url{https://doi.org/10.1109/AVSS.2018.8639163}
\showDOI{\tempurl}


\bibitem[He et~al\mbox{.}(2016)]%
        {he2016deep}
\bibfield{author}{\bibinfo{person}{Kaiming He}, \bibinfo{person}{Xiangyu
  Zhang}, \bibinfo{person}{Shaoqing Ren}, {and} \bibinfo{person}{Jian Sun}.}
  \bibinfo{year}{2016}\natexlab{}.
\newblock \showarticletitle{Deep residual learning for image recognition}. In
  \bibinfo{booktitle}{\emph{Proceedings of the IEEE conference on computer
  vision and pattern recognition}}. \bibinfo{pages}{770--778}.
\newblock


\bibitem[He et~al\mbox{.}(2019)]%
        {he_attgan_2018}
\bibfield{author}{\bibinfo{person}{Zhenliang He}, \bibinfo{person}{Wangmeng
  Zuo}, \bibinfo{person}{Meina Kan}, \bibinfo{person}{Shiguang Shan}, {and}
  \bibinfo{person}{Xilin Chen}.} \bibinfo{year}{2019}\natexlab{}.
\newblock \showarticletitle{Attgan: Facial attribute editing by only changing
  what you want}.
\newblock \bibinfo{journal}{\emph{IEEE transactions on image processing}}
  \bibinfo{volume}{28}, \bibinfo{number}{11} (\bibinfo{year}{2019}),
  \bibinfo{pages}{5464--5478}.
\newblock


\bibitem[Huang et~al\mbox{.}(2021)]%
        {huang_deepfake_2021}
\bibfield{author}{\bibinfo{person}{Jiajun Huang}, \bibinfo{person}{Xueyu Wang},
  \bibinfo{person}{Bo Du}, \bibinfo{person}{Pei Du}, {and}
  \bibinfo{person}{Chang Xu}.} \bibinfo{year}{2021}\natexlab{}.
\newblock \showarticletitle{DeepFake MNIST+: A DeepFake Facial Animation
  Dataset}. In \bibinfo{booktitle}{\emph{Proceedings of the IEEE/CVF
  International Conference on Computer Vision}}. \bibinfo{pages}{1973--1982}.
\newblock


\bibitem[Jagielski et~al\mbox{.}(2020)]%
        {jagielski2020high}
\bibfield{author}{\bibinfo{person}{Matthew Jagielski},
  \bibinfo{person}{Nicholas Carlini}, \bibinfo{person}{David Berthelot},
  \bibinfo{person}{Alex Kurakin}, {and} \bibinfo{person}{Nicolas Papernot}.}
  \bibinfo{year}{2020}\natexlab{}.
\newblock \showarticletitle{High accuracy and high fidelity extraction of
  neural networks}. In \bibinfo{booktitle}{\emph{29th USENIX Security Symposium
  (USENIX Security 20)}}. \bibinfo{pages}{1345--1362}.
\newblock


\bibitem[Jain and Korshunov(2021)]%
        {jain_improving_2021}
\bibfield{author}{\bibinfo{person}{Anubhav Jain} {and} \bibinfo{person}{Pavel
  Korshunov}.} \bibinfo{year}{2021}\natexlab{}.
\newblock \showarticletitle{Improving {Generalization} of {Deepfake}
  {Detection} by {Training} for {Attribution}}.
\newblock  (\bibinfo{year}{2021}).
\newblock
\newblock
\shownote{ISBN: 9781665432887}.


\bibitem[Karras et~al\mbox{.}(2021)]%
        {karras_alias-free_2021}
\bibfield{author}{\bibinfo{person}{Tero Karras}, \bibinfo{person}{Miika
  Aittala}, \bibinfo{person}{Samuli Laine}, \bibinfo{person}{Erik
  H{\"a}rk{\"o}nen}, \bibinfo{person}{Janne Hellsten}, \bibinfo{person}{Jaakko
  Lehtinen}, {and} \bibinfo{person}{Timo Aila}.}
  \bibinfo{year}{2021}\natexlab{}.
\newblock \showarticletitle{Alias-free generative adversarial networks}.
\newblock \bibinfo{journal}{\emph{Advances in Neural Information Processing
  Systems}}  \bibinfo{volume}{34} (\bibinfo{year}{2021}).
\newblock


\bibitem[Karras et~al\mbox{.}(2020)]%
        {karras_analyzing_2020}
\bibfield{author}{\bibinfo{person}{Tero Karras}, \bibinfo{person}{Samuli
  Laine}, \bibinfo{person}{Miika Aittala}, \bibinfo{person}{Janne Hellsten},
  \bibinfo{person}{Jaakko Lehtinen}, {and} \bibinfo{person}{Timo Aila}.}
  \bibinfo{year}{2020}\natexlab{}.
\newblock \showarticletitle{Analyzing and {Improving} the {Image} {Quality} of
  {StyleGAN}}. In \bibinfo{booktitle}{\emph{2020 {IEEE}/{CVF} {Conference} on
  {Computer} {Vision} and {Pattern} {Recognition} ({CVPR})}}.
  \bibinfo{pages}{8107--8116}.
\newblock
\urldef\tempurl%
\url{https://doi.org/10.1109/CVPR42600.2020.00813}
\showDOI{\tempurl}
\newblock
\shownote{ISSN: 2575-7075}.


\bibitem[Khalil et~al\mbox{.}(2021)]%
        {khalil_multi-layer_2021}
\bibfield{author}{\bibinfo{person}{Samar~Samir Khalil},
  \bibinfo{person}{Sherin~M. Youssef}, {and} \bibinfo{person}{Sherine~Nagy
  Saleh}.} \bibinfo{year}{2021}\natexlab{}.
\newblock \showarticletitle{A {Multi}-{Layer} {Capsule}-based {Forensics}
  {Model} for {Fake} {Detection} of {Digital} {Visual} {Media}}. In
  \bibinfo{booktitle}{\emph{2020 {International} {Conference} on
  {Communications}, {Signal} {Processing}, and their {Applications}
  ({ICCSPA})}}. \bibinfo{pages}{1--6}.
\newblock
\urldef\tempurl%
\url{https://doi.org/10.1109/ICCSPA49915.2021.9385719}
\showDOI{\tempurl}


\bibitem[Khan and Dai(2021)]%
        {khan_video_2021}
\bibfield{author}{\bibinfo{person}{Sohail~Ahmed Khan} {and}
  \bibinfo{person}{Hang Dai}.} \bibinfo{year}{2021}\natexlab{}.
\newblock \showarticletitle{Video Transformer for Deepfake Detection with
  Incremental Learning}. In \bibinfo{booktitle}{\emph{Proceedings of the 29th
  ACM International Conference on Multimedia}}. \bibinfo{pages}{1821--1828}.
\newblock


\bibitem[Khodabakhsh et~al\mbox{.}(2018)]%
        {khodabakhsh_fake_2018}
\bibfield{author}{\bibinfo{person}{Ali Khodabakhsh},
  \bibinfo{person}{Raghavendra Ramachandra}, \bibinfo{person}{Kiran Raja},
  \bibinfo{person}{Pankaj Wasnik}, {and} \bibinfo{person}{Christoph Busch}.}
  \bibinfo{year}{2018}\natexlab{}.
\newblock \showarticletitle{Fake {Face} {Detection} {Methods}: {Can} {They}
  {Be} {Generalized}?}. In \bibinfo{booktitle}{\emph{2018 {International}
  {Conference} of the {Biometrics} {Special} {Interest} {Group}, {BIOSIG}
  2018}}.
\newblock
\showISBNx{978-3-88579-676-3}
\urldef\tempurl%
\url{https://doi.org/10.23919/BIOSIG.2018.8553251}
\showDOI{\tempurl}


\bibitem[Khormali and Yuan(2021)]%
        {khormali_add_2021}
\bibfield{author}{\bibinfo{person}{Aminollah Khormali} {and}
  \bibinfo{person}{Jiann~Shiun Yuan}.} \bibinfo{year}{2021}\natexlab{}.
\newblock \showarticletitle{Add: {Attention}-based deepfake detection
  approach}.
\newblock \bibinfo{journal}{\emph{Big Data and Cognitive Computing}}
  \bibinfo{volume}{5}, \bibinfo{number}{4} (\bibinfo{year}{2021}).
\newblock
\showISSN{25042289}
\urldef\tempurl%
\url{https://doi.org/10.3390/bdcc5040049}
\showDOI{\tempurl}


\bibitem[Kim et~al\mbox{.}(2021a)]%
        {kim_cored_2021}
\bibfield{author}{\bibinfo{person}{Minha Kim}, \bibinfo{person}{Shahroz Tariq},
  {and} \bibinfo{person}{Simon~S. Woo}.} \bibinfo{year}{2021}\natexlab{a}.
\newblock \bibinfo{booktitle}{\emph{{CoReD}: {Generalizing} {Fake} {Media}
  {Detection} with {Continual} {Representation} using {Distillation}}}.
  Vol.~\bibinfo{volume}{1}.
\newblock \bibinfo{publisher}{Association for Computing Machinery}.
\newblock
\urldef\tempurl%
\url{https://doi.org/10.1145/3474085.3475535}
\showDOI{\tempurl}


\bibitem[Kim et~al\mbox{.}(2021b)]%
        {kim_fretal_2021}
\bibfield{author}{\bibinfo{person}{Minha Kim}, \bibinfo{person}{Shahroz Tariq},
  {and} \bibinfo{person}{Simon~S Woo}.} \bibinfo{year}{2021}\natexlab{b}.
\newblock \showarticletitle{Fretal: Generalizing deepfake detection using
  knowledge distillation and representation learning}. In
  \bibinfo{booktitle}{\emph{Proceedings of the IEEE/CVF Conference on Computer
  Vision and Pattern Recognition}}. \bibinfo{pages}{1001--1012}.
\newblock


\bibitem[Kingma and Ba(2015)]%
        {kingma_adam_2017}
\bibfield{author}{\bibinfo{person}{Diederik~P. Kingma} {and}
  \bibinfo{person}{Jimmy Ba}.} \bibinfo{year}{2015}\natexlab{}.
\newblock \showarticletitle{Adam: {A} Method for Stochastic Optimization}. In
  \bibinfo{booktitle}{\emph{3rd International Conference on Learning
  Representations, {ICLR} 2015, San Diego, CA, USA, May 7-9, 2015, Conference
  Track Proceedings}}, \bibfield{editor}{\bibinfo{person}{Yoshua Bengio} {and}
  \bibinfo{person}{Yann LeCun}} (Eds.).
\newblock


\bibitem[Kordopatis-Zilos et~al\mbox{.}(2019)]%
        {kordopatis2019visil}
\bibfield{author}{\bibinfo{person}{Giorgos Kordopatis-Zilos},
  \bibinfo{person}{Symeon Papadopoulos}, \bibinfo{person}{Ioannis Patras},
  {and} \bibinfo{person}{Ioannis Kompatsiaris}.}
  \bibinfo{year}{2019}\natexlab{}.
\newblock \showarticletitle{Visil: Fine-grained spatio-temporal video
  similarity learning}. In \bibinfo{booktitle}{\emph{Proceedings of the
  IEEE/CVF International Conference on Computer Vision}}.
  \bibinfo{pages}{6351--6360}.
\newblock


\bibitem[Kordopatis{-}Zilos et~al\mbox{.}(2021)]%
        {kordopatis-zilos_dns_2021}
\bibfield{author}{\bibinfo{person}{Giorgos Kordopatis{-}Zilos},
  \bibinfo{person}{Christos Tzelepis}, \bibinfo{person}{Symeon Papadopoulos},
  \bibinfo{person}{Ioannis Kompatsiaris}, {and} \bibinfo{person}{Ioannis
  Patras}.} \bibinfo{year}{2021}\natexlab{}.
\newblock \showarticletitle{DnS: Distill-and-Select for Efficient and Accurate
  Video Indexing and Retrieval}.
\newblock \bibinfo{journal}{\emph{CoRR}}  \bibinfo{volume}{abs/2106.13266}
  (\bibinfo{year}{2021}).
\newblock
\showeprint[arXiv]{2106.13266}


\bibitem[Korshunov and Marcel(2018)]%
        {korshunov_deepfakes_2018}
\bibfield{author}{\bibinfo{person}{Pavel Korshunov} {and}
  \bibinfo{person}{Sebastien Marcel}.} \bibinfo{year}{2018}\natexlab{}.
\newblock \showarticletitle{{DeepFakes}: a {New} {Threat} to {Face}
  {Recognition}? {Assessment} and {Detection}}.
\newblock  (\bibinfo{year}{2018}), \bibinfo{pages}{1--5}.
\newblock
\newblock
\shownote{arXiv: 1812.08685}.


\bibitem[Kwon et~al\mbox{.}(2021)]%
        {kwon_kodf_2021}
\bibfield{author}{\bibinfo{person}{Patrick Kwon}, \bibinfo{person}{Jaeseong
  You}, \bibinfo{person}{Gyuhyeon Nam}, \bibinfo{person}{Sungwoo Park}, {and}
  \bibinfo{person}{Gyeongsu Chae}.} \bibinfo{year}{2021}\natexlab{}.
\newblock \showarticletitle{Kodf: A large-scale korean deepfake detection
  dataset}. In \bibinfo{booktitle}{\emph{Proceedings of the IEEE/CVF
  International Conference on Computer Vision}}. \bibinfo{pages}{10744--10753}.
\newblock


\bibitem[Le et~al\mbox{.}(2022)]%
        {le_robust_2022}
\bibfield{author}{\bibinfo{person}{Trung{-}Nghia Le}, \bibinfo{person}{Huy~H.
  Nguyen}, \bibinfo{person}{Junichi Yamagishi}, {and} \bibinfo{person}{Isao
  Echizen}.} \bibinfo{year}{2022}\natexlab{}.
\newblock \showarticletitle{Robust Deepfake On Unrestricted Media: Generation
  And Detection}.
\newblock \bibinfo{journal}{\emph{CoRR}}  \bibinfo{volume}{abs/2202.06228}
  (\bibinfo{year}{2022}).
\newblock
\showeprint[arXiv]{2202.06228}


\bibitem[Le et~al\mbox{.}(2021)]%
        {le_openforensics_2021}
\bibfield{author}{\bibinfo{person}{Trung-Nghia Le}, \bibinfo{person}{Huy~H
  Nguyen}, \bibinfo{person}{Junichi Yamagishi}, {and} \bibinfo{person}{Isao
  Echizen}.} \bibinfo{year}{2021}\natexlab{}.
\newblock \showarticletitle{Openforensics: Large-scale challenging dataset for
  multi-face forgery detection and segmentation in-the-wild}. In
  \bibinfo{booktitle}{\emph{Proceedings of the IEEE/CVF International
  Conference on Computer Vision}}. \bibinfo{pages}{10117--10127}.
\newblock


\bibitem[Li et~al\mbox{.}(2020a)]%
        {li_face_2020}
\bibfield{author}{\bibinfo{person}{Lingzhi Li}, \bibinfo{person}{Jianmin Bao},
  \bibinfo{person}{Ting Zhang}, \bibinfo{person}{Hao Yang},
  \bibinfo{person}{Dong Chen}, \bibinfo{person}{Fang Wen}, {and}
  \bibinfo{person}{Baining Guo}.} \bibinfo{year}{2020}\natexlab{a}.
\newblock \showarticletitle{Face {X}-ray for more general face forgery
  detection}.
\newblock \bibinfo{journal}{\emph{Proceedings of the IEEE Computer Society
  Conference on Computer Vision and Pattern Recognition}}
  (\bibinfo{year}{2020}), \bibinfo{pages}{5000--5009}.
\newblock
\showISSN{10636919}
\urldef\tempurl%
\url{https://doi.org/10.1109/CVPR42600.2020.00505}
\showDOI{\tempurl}


\bibitem[Li and Lyu(2019)]%
        {li_exposing_2018}
\bibfield{author}{\bibinfo{person}{Yuezun Li} {and} \bibinfo{person}{Siwei
  Lyu}.} \bibinfo{year}{2019}\natexlab{}.
\newblock \showarticletitle{Exposing DeepFake Videos By Detecting Face Warping
  Artifacts}. In \bibinfo{booktitle}{\emph{IEEE Conference on Computer Vision
  and Pattern Recognition Workshops (CVPRW)}}.
\newblock


\bibitem[Li et~al\mbox{.}(2020b)]%
        {li_celeb-df_2020}
\bibfield{author}{\bibinfo{person}{Yuezun Li}, \bibinfo{person}{Xin Yang},
  \bibinfo{person}{Pu Sun}, \bibinfo{person}{Honggang Qi}, {and}
  \bibinfo{person}{Siwei Lyu}.} \bibinfo{year}{2020}\natexlab{b}.
\newblock \showarticletitle{Celeb-{DF}: {A} {Large}-{Scale} {Challenging}
  {Dataset} for {DeepFake} {Forensics}}.
\newblock \bibinfo{journal}{\emph{Proceedings of the IEEE Computer Society
  Conference on Computer Vision and Pattern Recognition}}
  (\bibinfo{year}{2020}), \bibinfo{pages}{3204--3213}.
\newblock
\showISSN{10636919}
\urldef\tempurl%
\url{https://doi.org/10.1109/CVPR42600.2020.00327}
\showDOI{\tempurl}


\bibitem[Li et~al\mbox{.}(2021)]%
        {li_deepfake-o-meter_2021}
\bibfield{author}{\bibinfo{person}{Yuezun Li}, \bibinfo{person}{Cong Zhang},
  \bibinfo{person}{Pu Sun}, \bibinfo{person}{Lipeng Ke}, \bibinfo{person}{Yan
  Ju}, \bibinfo{person}{Honggang Qi}, {and} \bibinfo{person}{Siwei Lyu}.}
  \bibinfo{year}{2021}\natexlab{}.
\newblock \showarticletitle{DeepFake-o-meter: An Open Platform for DeepFake
  Detection}. In \bibinfo{booktitle}{\emph{2021 IEEE Security and Privacy
  Workshops (SPW)}}. IEEE, \bibinfo{pages}{277--281}.
\newblock


\bibitem[Madry et~al\mbox{.}(2018)]%
        {madry_towards_2019}
\bibfield{author}{\bibinfo{person}{Aleksander Madry},
  \bibinfo{person}{Aleksandar Makelov}, \bibinfo{person}{Ludwig Schmidt},
  \bibinfo{person}{Dimitris Tsipras}, {and} \bibinfo{person}{Adrian Vladu}.}
  \bibinfo{year}{2018}\natexlab{}.
\newblock \showarticletitle{Towards Deep Learning Models Resistant to
  Adversarial Attacks}. In \bibinfo{booktitle}{\emph{International Conference
  on Learning Representations}}.
\newblock
\urldef\tempurl%
\url{https://openreview.net/forum?id=rJzIBfZAb}
\showURL{%
\tempurl}


\bibitem[Malik et~al\mbox{.}(2022)]%
        {malik_deepfake_2022}
\bibfield{author}{\bibinfo{person}{Asad Malik}, \bibinfo{person}{Minoru
  Kuribayashi}, \bibinfo{person}{Sani~M. Abdullahi}, {and}
  \bibinfo{person}{Ahmad~Neyaz Khan}.} \bibinfo{year}{2022}\natexlab{}.
\newblock \showarticletitle{{DeepFake} {Detection} for {Human} {Face} {Images}
  and {Videos}: {A} {Survey}}.
\newblock \bibinfo{journal}{\emph{IEEE Access}} (\bibinfo{year}{2022}),
  \bibinfo{pages}{1--1}.
\newblock
\showISSN{2169-3536}
\urldef\tempurl%
\url{https://doi.org/10.1109/ACCESS.2022.3151186}
\showDOI{\tempurl}
\newblock
\shownote{Conference Name: IEEE Access}.


\bibitem[Masood et~al\mbox{.}(2021)]%
        {masood_deepfakes_2021}
\bibfield{author}{\bibinfo{person}{Momina Masood}, \bibinfo{person}{Marriam
  Nawaz}, \bibinfo{person}{Khalid~Mahmood Malik}, \bibinfo{person}{Ali Javed},
  {and} \bibinfo{person}{Aun Irtaza}.} \bibinfo{year}{2021}\natexlab{}.
\newblock \showarticletitle{Deepfakes {Generation} and {Detection}:
  {State}-of-the-art, open challenges, countermeasures, and way forward}.
\newblock  (\bibinfo{year}{2021}).
\newblock
\newblock
\shownote{arXiv: 2103.00484}.


\bibitem[Mehra et~al\mbox{.}(2021)]%
        {mehra_deepfake_2021}
\bibfield{author}{\bibinfo{person}{Akul Mehra}, \bibinfo{person}{Luuk
  Spreeuwers}, {and} \bibinfo{person}{Nicola Strisciuglio}.}
  \bibinfo{year}{2021}\natexlab{}.
\newblock \showarticletitle{Deepfake {Detection} using {Capsule} {Networks} and
  {Long} {Short}-{Term} {Memory} {Networks}:}. In
  \bibinfo{booktitle}{\emph{Proceedings of the 16th {International} {Joint}
  {Conference} on {Computer} {Vision}, {Imaging} and {Computer} {Graphics}
  {Theory} and {Applications}}}. \bibinfo{publisher}{SCITEPRESS - Science and
  Technology Publications}, \bibinfo{address}{Online Streaming, --- Select a
  Country ---}, \bibinfo{pages}{407--414}.
\newblock
\showISBNx{978-989-758-488-6}
\urldef\tempurl%
\url{https://doi.org/10.5220/0010289004070414}
\showDOI{\tempurl}


\bibitem[MinIO(2022)]%
        {minio}
\bibfield{author}{\bibinfo{person}{MinIO}.} \bibinfo{year}{2022}\natexlab{}.
\newblock \bibinfo{title}{MinIO}.
\newblock \bibinfo{howpublished}{\url{https://min.io}}.
\newblock


\bibitem[Mirsky and Lee(2021)]%
        {mirsky_creation_2021}
\bibfield{author}{\bibinfo{person}{Yisroel Mirsky} {and} \bibinfo{person}{Wenke
  Lee}.} \bibinfo{year}{2021}\natexlab{}.
\newblock \showarticletitle{The {Creation} and {Detection} of {Deepfakes}}.
\newblock \bibinfo{journal}{\emph{Comput. Surveys}} \bibinfo{volume}{54},
  \bibinfo{number}{1} (\bibinfo{year}{2021}), \bibinfo{pages}{1--41}.
\newblock
\showISSN{15577341}
\urldef\tempurl%
\url{https://doi.org/10.1145/3425780}
\showDOI{\tempurl}


\bibitem[Mitchell et~al\mbox{.}(2019)]%
        {mitchell_model_2019}
\bibfield{author}{\bibinfo{person}{Margaret Mitchell}, \bibinfo{person}{Simone
  Wu}, \bibinfo{person}{Andrew Zaldivar}, \bibinfo{person}{Parker Barnes},
  \bibinfo{person}{Lucy Vasserman}, \bibinfo{person}{Ben Hutchinson},
  \bibinfo{person}{Elena Spitzer}, \bibinfo{person}{Inioluwa~Deborah Raji},
  {and} \bibinfo{person}{Timnit Gebru}.} \bibinfo{year}{2019}\natexlab{}.
\newblock \showarticletitle{Model {Cards} for {Model} {Reporting}}.
\newblock \bibinfo{journal}{\emph{Proceedings of the Conference on Fairness,
  Accountability, and Transparency}} (\bibinfo{date}{Jan.}
  \bibinfo{year}{2019}), \bibinfo{pages}{220--229}.
\newblock
\urldef\tempurl%
\url{https://doi.org/10.1145/3287560.3287596}
\showDOI{\tempurl}


\bibitem[Nguyen et~al\mbox{.}(2019)]%
        {nguyen_capsule-forensics_2019}
\bibfield{author}{\bibinfo{person}{Huy~H. Nguyen}, \bibinfo{person}{Junichi
  Yamagishi}, {and} \bibinfo{person}{Isao Echizen}.}
  \bibinfo{year}{2019}\natexlab{}.
\newblock \showarticletitle{Capsule-forensics: {Using} {Capsule} {Networks} to
  {Detect} {Forged} {Images} and {Videos}}.
\newblock \bibinfo{journal}{\emph{ICASSP, IEEE International Conference on
  Acoustics, Speech and Signal Processing - Proceedings}}
  \bibinfo{volume}{2019-May} (\bibinfo{year}{2019}),
  \bibinfo{pages}{2307--2311}.
\newblock
\showISSN{15206149}
\urldef\tempurl%
\url{https://doi.org/10.1109/ICASSP.2019.8682602}
\showDOI{\tempurl}
\newblock
\shownote{ISBN: 9781479981311}.


\bibitem[Nguyen et~al\mbox{.}(2022)]%
        {nguyen_capsule-forensics_2022}
\bibfield{author}{\bibinfo{person}{Huy~H. Nguyen}, \bibinfo{person}{Junichi
  Yamagishi}, {and} \bibinfo{person}{Isao Echizen}.}
  \bibinfo{year}{2022}\natexlab{}.
\newblock \showarticletitle{Capsule-{Forensics} {Networks} for {Deepfake}
  {Detection}}.
\newblock In \bibinfo{booktitle}{\emph{Handbook of {Digital} {Face}
  {Manipulation} and {Detection}: {From} {DeepFakes} to {Morphing} {Attacks}}},
  \bibfield{editor}{\bibinfo{person}{Christian Rathgeb}, \bibinfo{person}{Ruben
  Tolosana}, \bibinfo{person}{Ruben Vera-Rodriguez}, {and}
  \bibinfo{person}{Christoph Busch}} (Eds.). \bibinfo{publisher}{Springer
  International Publishing}, \bibinfo{address}{Cham},
  \bibinfo{pages}{275--301}.
\newblock
\showISBNx{978-3-030-87664-7}
\urldef\tempurl%
\url{https://doi.org/10.1007/978-3-030-87664-7_13}
\showDOI{\tempurl}


\bibitem[NVIDIA(2022)]%
        {triton}
\bibfield{author}{\bibinfo{person}{NVIDIA}.} \bibinfo{year}{2022}\natexlab{}.
\newblock \bibinfo{title}{NVIDIA TRITON INFERENCE SERVER}.
\newblock
  \bibinfo{howpublished}{\url{https://github.com/triton-inference-server/server}}.
\newblock


\bibitem[OpenCV-team(2022)]%
        {opencv}
\bibfield{author}{\bibinfo{person}{OpenCV-team}.}
  \bibinfo{year}{2022}\natexlab{}.
\newblock \bibinfo{title}{OpenCV}.
\newblock \bibinfo{howpublished}{\url{https://opencv.org}}.
\newblock


\bibitem[Passos et~al\mbox{.}(2022)]%
        {passos_review_2022}
\bibfield{author}{\bibinfo{person}{Leandro~A. Passos}, \bibinfo{person}{Danilo
  Jodas}, \bibinfo{person}{Kelton A.~P. da Costa}, \bibinfo{person}{Luis
  A.~Souza Júnior}, \bibinfo{person}{Danilo Colombo}, {and}
  \bibinfo{person}{João~Paulo Papa}.} \bibinfo{year}{2022}\natexlab{}.
\newblock \showarticletitle{A {Review} of {Deep} {Learning}-based {Approaches}
  for {Deepfake} {Content} {Detection}}.
\newblock \bibinfo{journal}{\emph{arXiv:2202.06095 [cs]}} (\bibinfo{date}{Feb.}
  \bibinfo{year}{2022}).
\newblock


\bibitem[Pu et~al\mbox{.}(2021)]%
        {pu_deepfake_2021}
\bibfield{author}{\bibinfo{person}{Jiameng Pu}, \bibinfo{person}{Neal
  Mangaokar}, \bibinfo{person}{Lauren Kelly}, \bibinfo{person}{Parantapa
  Bhattacharya}, \bibinfo{person}{Kavya Sundaram}, \bibinfo{person}{Mobin
  Javed}, \bibinfo{person}{Bolun Wang}, {and} \bibinfo{person}{Bimal
  Viswanath}.} \bibinfo{year}{2021}\natexlab{}.
\newblock \bibinfo{booktitle}{\emph{Deepfake videos in the wild: {Analysis} and
  detection}}. Vol.~\bibinfo{volume}{1}.
\newblock \bibinfo{publisher}{Association for Computing Machinery}.
\newblock
\showISBNx{978-1-4503-8312-7}
\urldef\tempurl%
\url{https://doi.org/10.1145/3442381.3449978}
\showDOI{\tempurl}
\newblock
\shownote{Publication Title: The Web Conference 2021 - Proceedings of the World
  Wide Web Conference, WWW 2021 Issue: 1}.


\bibitem[PyTorch(2022)]%
        {torchscript}
\bibfield{author}{\bibinfo{person}{PyTorch}.} \bibinfo{year}{2022}\natexlab{}.
\newblock \bibinfo{title}{TorchScript}.
\newblock
  \bibinfo{howpublished}{\url{https://pytorch.org/docs/stable/jit.html}}.
\newblock


\bibitem[Rao and Frtunikj(2018)]%
        {rao_deep_2018}
\bibfield{author}{\bibinfo{person}{Qing Rao} {and} \bibinfo{person}{Jelena
  Frtunikj}.} \bibinfo{year}{2018}\natexlab{}.
\newblock \showarticletitle{Deep learning for self-driving cars: chances and
  challenges}. In \bibinfo{booktitle}{\emph{Proceedings of the 1st
  {International} {Workshop} on {Software} {Engineering} for {AI} in
  {Autonomous} {Systems}}}. \bibinfo{publisher}{ACM},
  \bibinfo{address}{Gothenburg Sweden}, \bibinfo{pages}{35--38}.
\newblock
\showISBNx{978-1-4503-5739-5}
\urldef\tempurl%
\url{https://doi.org/10.1145/3194085.3194087}
\showDOI{\tempurl}


\bibitem[Redis(2021)]%
        {redis}
\bibfield{author}{\bibinfo{person}{Redis}.} \bibinfo{year}{2021}\natexlab{}.
\newblock \bibinfo{title}{Redis}.
\newblock \bibinfo{howpublished}{\url{https://redis.io/}}.
\newblock


\bibitem[Ridnik et~al\mbox{.}(2021)]%
        {ridnik_imagenet-21k_2021}
\bibfield{author}{\bibinfo{person}{Tal Ridnik}, \bibinfo{person}{Emanuel
  Ben-Baruch}, \bibinfo{person}{Asaf Noy}, {and} \bibinfo{person}{Lihi
  Zelnik-Manor}.} \bibinfo{year}{2021}\natexlab{}.
\newblock \showarticletitle{{ImageNet}-{21K} {Pretraining} for the {Masses}}.
\newblock \bibinfo{journal}{\emph{arXiv:2104.10972 [cs]}} (\bibinfo{date}{Aug.}
  \bibinfo{year}{2021}).
\newblock


\bibitem[Rossler et~al\mbox{.}(2019)]%
        {rossler_faceforensics_2019}
\bibfield{author}{\bibinfo{person}{Andreas Rossler}, \bibinfo{person}{Davide
  Cozzolino}, \bibinfo{person}{Luisa Verdoliva}, \bibinfo{person}{Christian
  Riess}, \bibinfo{person}{Justus Thies}, {and} \bibinfo{person}{Matthias
  Niessner}.} \bibinfo{year}{2019}\natexlab{}.
\newblock \showarticletitle{{FaceForensics}++: {Learning} to detect manipulated
  facial images}.
\newblock \bibinfo{journal}{\emph{Proceedings of the IEEE International
  Conference on Computer Vision}}  \bibinfo{volume}{2019-Octob}
  (\bibinfo{year}{2019}), \bibinfo{pages}{1--11}.
\newblock
\showISSN{15505499}
\urldef\tempurl%
\url{https://doi.org/10.1109/ICCV.2019.00009}
\showDOI{\tempurl}


\bibitem[Sabour et~al\mbox{.}(2017)]%
        {sabour2017dynamic}
\bibfield{author}{\bibinfo{person}{Sara Sabour}, \bibinfo{person}{Nicholas
  Frosst}, {and} \bibinfo{person}{Geoffrey~E Hinton}.}
  \bibinfo{year}{2017}\natexlab{}.
\newblock \showarticletitle{Dynamic routing between capsules}.
\newblock \bibinfo{journal}{\emph{Advances in neural information processing
  systems}}  \bibinfo{volume}{30} (\bibinfo{year}{2017}).
\newblock


\bibitem[Shen et~al\mbox{.}(2020)]%
        {shen_interpreting_2020}
\bibfield{author}{\bibinfo{person}{Yujun Shen}, \bibinfo{person}{Jinjin Gu},
  \bibinfo{person}{Xiaoou Tang}, {and} \bibinfo{person}{Bolei Zhou}.}
  \bibinfo{year}{2020}\natexlab{}.
\newblock \showarticletitle{Interpreting the latent space of gans for semantic
  face editing}. In \bibinfo{booktitle}{\emph{Proceedings of the IEEE/CVF
  Conference on Computer Vision and Pattern Recognition}}.
  \bibinfo{pages}{9243--9252}.
\newblock


\bibitem[Sun et~al\mbox{.}(2021)]%
        {sun_faketransformer_2021}
\bibfield{author}{\bibinfo{person}{Yuyang Sun}, \bibinfo{person}{Zhiyong
  Zhang}, \bibinfo{person}{Changzhen Qiu}, \bibinfo{person}{Liang Wang}, {and}
  \bibinfo{person}{Zekai Wang}.} \bibinfo{year}{2021}\natexlab{}.
\newblock \showarticletitle{{FakeTransformer}: {Exposing} {Face} {Forgery}
  {From} {Spatial}-{Temporal} {Representation} {Modeled} {By} {Facial} {Pixel}
  {Variations}}.
\newblock  (\bibinfo{year}{2021}).
\newblock
\newblock
\shownote{arXiv: 2111.07601}.


\bibitem[Suwajanakorn et~al\mbox{.}(2017)]%
        {suwajanakorn_synthesizing_2017}
\bibfield{author}{\bibinfo{person}{Supasorn Suwajanakorn},
  \bibinfo{person}{Steven~M. Seitz}, {and} \bibinfo{person}{Ira
  Kemelmacher-Shlizerman}.} \bibinfo{year}{2017}\natexlab{}.
\newblock \showarticletitle{Synthesizing {Obama}: learning lip sync from
  audio}.
\newblock \bibinfo{journal}{\emph{ACM Trans. Graph.}} \bibinfo{volume}{36},
  \bibinfo{number}{4} (\bibinfo{date}{July} \bibinfo{year}{2017}),
  \bibinfo{pages}{1--13}.
\newblock
\showISSN{0730-0301, 1557-7368}
\urldef\tempurl%
\url{https://doi.org/10.1145/3072959.3073640}
\showDOI{\tempurl}


\bibitem[Tan and Le(2019)]%
        {tan_efficientnet_2020}
\bibfield{author}{\bibinfo{person}{Mingxing Tan} {and} \bibinfo{person}{Quoc
  Le}.} \bibinfo{year}{2019}\natexlab{}.
\newblock \showarticletitle{Efficientnet: Rethinking model scaling for
  convolutional neural networks}. In \bibinfo{booktitle}{\emph{International
  conference on machine learning}}. PMLR, \bibinfo{pages}{6105--6114}.
\newblock


\bibitem[Tan and Le(2021)]%
        {tan_efficientnetv2_2021}
\bibfield{author}{\bibinfo{person}{Mingxing Tan} {and} \bibinfo{person}{Quoc
  Le}.} \bibinfo{year}{2021}\natexlab{}.
\newblock \showarticletitle{Efficientnetv2: Smaller models and faster
  training}. In \bibinfo{booktitle}{\emph{International Conference on Machine
  Learning}}. PMLR, \bibinfo{pages}{10096--10106}.
\newblock


\bibitem[Tariq et~al\mbox{.}(2021)]%
        {tariq_one_2021}
\bibfield{author}{\bibinfo{person}{Shahroz Tariq}, \bibinfo{person}{Sangyup
  Lee}, {and} \bibinfo{person}{Simon Woo}.} \bibinfo{year}{2021}\natexlab{}.
\newblock \showarticletitle{One detector to rule them all: {Towards} a general
  deepfake attack detection framework}.
\newblock \bibinfo{journal}{\emph{The Web Conference 2021 - Proceedings of the
  World Wide Web Conference, WWW 2021}} (\bibinfo{year}{2021}),
  \bibinfo{pages}{3625--3637}.
\newblock
\urldef\tempurl%
\url{https://doi.org/10.1145/3442381.3449809}
\showDOI{\tempurl}
\newblock
\shownote{ISBN: 9781450383127}.


\bibitem[Teyssou et~al\mbox{.}(2017)]%
        {teyssou2017invid}
\bibfield{author}{\bibinfo{person}{Denis Teyssou}, \bibinfo{person}{Jean-Michel
  Leung}, \bibinfo{person}{Evlampios Apostolidis},
  \bibinfo{person}{Konstantinos Apostolidis}, \bibinfo{person}{Symeon
  Papadopoulos}, \bibinfo{person}{Markos Zampoglou}, \bibinfo{person}{Olga
  Papadopoulou}, {and} \bibinfo{person}{Vasileios Mezaris}.}
  \bibinfo{year}{2017}\natexlab{}.
\newblock \showarticletitle{The InVID plug-in: web video verification on the
  browser}. In \bibinfo{booktitle}{\emph{Proceedings of the first international
  workshop on multimedia verification}}. \bibinfo{pages}{23--30}.
\newblock


\bibitem[Thies et~al\mbox{.}(2016)]%
        {thies_face2face_2020}
\bibfield{author}{\bibinfo{person}{Justus Thies}, \bibinfo{person}{Michael
  Zollhofer}, \bibinfo{person}{Marc Stamminger}, \bibinfo{person}{Christian
  Theobalt}, {and} \bibinfo{person}{Matthias Nie{\ss}ner}.}
  \bibinfo{year}{2016}\natexlab{}.
\newblock \showarticletitle{Face2face: Real-time face capture and reenactment
  of rgb videos}. In \bibinfo{booktitle}{\emph{Proceedings of the IEEE
  conference on computer vision and pattern recognition}}.
  \bibinfo{pages}{2387--2395}.
\newblock


\bibitem[tiangolo(2022)]%
        {fastapi}
\bibfield{author}{\bibinfo{person}{tiangolo}.} \bibinfo{year}{2022}\natexlab{}.
\newblock \bibinfo{title}{FastAPI}.
\newblock \bibinfo{howpublished}{\url{https://fastapi.tiangolo.com/}}.
\newblock


\bibitem[Tolias et~al\mbox{.}(2016)]%
        {tolias2016}
\bibfield{author}{\bibinfo{person}{Giorgos Tolias}, \bibinfo{person}{Ronan
  Sicre}, {and} \bibinfo{person}{Herv{\'e} J{\'e}gou}.}
  \bibinfo{year}{2016}\natexlab{}.
\newblock \showarticletitle{Particular object retrieval with integral
  max-pooling of CNN activations}. In \bibinfo{booktitle}{\emph{Proceedings of
  the International Conference on Learning Representations}}.
\newblock


\bibitem[Tran et~al\mbox{.}(2021)]%
        {tran_high_2021}
\bibfield{author}{\bibinfo{person}{Van~Nhan Tran}, \bibinfo{person}{Suk~Hwan
  Lee}, \bibinfo{person}{Hoanh~Su Le}, {and} \bibinfo{person}{Ki~Ryong Kwon}.}
  \bibinfo{year}{2021}\natexlab{}.
\newblock \showarticletitle{High performance deepfake video detection on
  cnn-based with attention target-specific regions and manual distillation
  extraction}.
\newblock \bibinfo{journal}{\emph{Applied Sciences (Switzerland)}}
  \bibinfo{volume}{11}, \bibinfo{number}{16} (\bibinfo{year}{2021}).
\newblock
\showISSN{20763417}
\urldef\tempurl%
\url{https://doi.org/10.3390/app11167678}
\showDOI{\tempurl}


\bibitem[Vaswani et~al\mbox{.}(2017)]%
        {vaswani2017attention}
\bibfield{author}{\bibinfo{person}{Ashish Vaswani}, \bibinfo{person}{Noam
  Shazeer}, \bibinfo{person}{Niki Parmar}, \bibinfo{person}{Jakob Uszkoreit},
  \bibinfo{person}{Llion Jones}, \bibinfo{person}{Aidan~N Gomez},
  \bibinfo{person}{{\L}ukasz Kaiser}, {and} \bibinfo{person}{Illia
  Polosukhin}.} \bibinfo{year}{2017}\natexlab{}.
\newblock \showarticletitle{Attention is all you need}.
\newblock \bibinfo{journal}{\emph{Advances in neural information processing
  systems}}  \bibinfo{volume}{30} (\bibinfo{year}{2017}).
\newblock


\bibitem[Xu et~al\mbox{.}(2022)]%
        {xu_adversarial_2022}
\bibfield{author}{\bibinfo{person}{Ying Xu}, \bibinfo{person}{Kiran Raja},
  \bibinfo{person}{Raghavendra Ramachandra}, {and} \bibinfo{person}{Christoph
  Busch}.} \bibinfo{year}{2022}\natexlab{}.
\newblock \showarticletitle{Adversarial {Attacks} on {Face} {Recognition}
  {Systems}}.
\newblock In \bibinfo{booktitle}{\emph{Handbook of {Digital} {Face}
  {Manipulation} and {Detection}: {From} {DeepFakes} to {Morphing} {Attacks}}},
  \bibfield{editor}{\bibinfo{person}{Christian Rathgeb}, \bibinfo{person}{Ruben
  Tolosana}, \bibinfo{person}{Ruben Vera-Rodriguez}, {and}
  \bibinfo{person}{Christoph Busch}} (Eds.). \bibinfo{publisher}{Springer
  International Publishing}, \bibinfo{address}{Cham},
  \bibinfo{pages}{139--161}.
\newblock
\showISBNx{978-3-030-87664-7}
\urldef\tempurl%
\url{https://doi.org/10.1007/978-3-030-87664-7_7}
\showDOI{\tempurl}


\bibitem[Yang et~al\mbox{.}(2019)]%
        {yang_exposing_2019}
\bibfield{author}{\bibinfo{person}{Xin Yang}, \bibinfo{person}{Yuezun Li},
  {and} \bibinfo{person}{Siwei Lyu}.} \bibinfo{year}{2019}\natexlab{}.
\newblock \showarticletitle{Exposing {Deep} {Fakes} {Using} {Inconsistent}
  {Head} {Poses}}.
\newblock \bibinfo{journal}{\emph{ICASSP, IEEE International Conference on
  Acoustics, Speech and Signal Processing - Proceedings}}
  \bibinfo{volume}{2019-May} (\bibinfo{year}{2019}),
  \bibinfo{pages}{8261--8265}.
\newblock
\showISSN{15206149}
\urldef\tempurl%
\url{https://doi.org/10.1109/ICASSP.2019.8683164}
\showDOI{\tempurl}
\newblock
\shownote{ISBN: 9781479981311}.


\bibitem[yt~dl(2022)]%
        {youtube-dl}
\bibfield{author}{\bibinfo{person}{yt dl}.} \bibinfo{year}{2022}\natexlab{}.
\newblock \bibinfo{title}{Youtube-DL}.
\newblock \bibinfo{howpublished}{\url{https://youtube-dl.org}}.
\newblock


\bibitem[yt~dlp(2022)]%
        {youtube-dlp}
\bibfield{author}{\bibinfo{person}{yt dlp}.} \bibinfo{year}{2022}\natexlab{}.
\newblock \bibinfo{title}{Youtube-DLP}.
\newblock \bibinfo{howpublished}{\url{https://github.com/yt-dlp/yt-dlp}}.
\newblock


\bibitem[Zhao et~al\mbox{.}(2021b)]%
        {zhao2021deep}
\bibfield{author}{\bibinfo{person}{Bo Zhao}, \bibinfo{person}{Shaozeng Zhang},
  \bibinfo{person}{Chunxue Xu}, \bibinfo{person}{Yifan Sun}, {and}
  \bibinfo{person}{Chengbin Deng}.} \bibinfo{year}{2021}\natexlab{b}.
\newblock \showarticletitle{Deep fake geography? When geospatial data encounter
  Artificial Intelligence}.
\newblock \bibinfo{journal}{\emph{Cartography and Geographic Information
  Science}} \bibinfo{volume}{48}, \bibinfo{number}{4} (\bibinfo{year}{2021}),
  \bibinfo{pages}{338--352}.
\newblock


\bibitem[Zhao et~al\mbox{.}(2021c)]%
        {zhao_multi-attentional_2021}
\bibfield{author}{\bibinfo{person}{Hanqing Zhao}, \bibinfo{person}{Wenbo Zhou},
  \bibinfo{person}{Dongdong Chen}, \bibinfo{person}{Tianyi Wei},
  \bibinfo{person}{Weiming Zhang}, {and} \bibinfo{person}{Nenghai Yu}.}
  \bibinfo{year}{2021}\natexlab{c}.
\newblock \showarticletitle{Multi-attentional deepfake detection}. In
  \bibinfo{booktitle}{\emph{Proceedings of the IEEE/CVF Conference on Computer
  Vision and Pattern Recognition}}. \bibinfo{pages}{2185--2194}.
\newblock


\bibitem[Zhao et~al\mbox{.}(2021a)]%
        {zhao_learning_2020}
\bibfield{author}{\bibinfo{person}{Tianchen Zhao}, \bibinfo{person}{Xiang Xu},
  \bibinfo{person}{Mingze Xu}, \bibinfo{person}{Hui Ding},
  \bibinfo{person}{Yuanjun Xiong}, {and} \bibinfo{person}{Wei Xia}.}
  \bibinfo{year}{2021}\natexlab{a}.
\newblock \showarticletitle{Learning self-consistency for deepfake detection}.
  In \bibinfo{booktitle}{\emph{Proceedings of the IEEE/CVF International
  Conference on Computer Vision}}. \bibinfo{pages}{15023--15033}.
\newblock


\bibitem[Zhou et~al\mbox{.}(2021)]%
        {zhou_face_2021}
\bibfield{author}{\bibinfo{person}{Tianfei Zhou}, \bibinfo{person}{Wenguan
  Wang}, \bibinfo{person}{Zhiyuan Liang}, {and} \bibinfo{person}{Jianbing
  Shen}.} \bibinfo{year}{2021}\natexlab{}.
\newblock \showarticletitle{Face forensics in the wild}. In
  \bibinfo{booktitle}{\emph{Proceedings of the IEEE/CVF Conference on Computer
  Vision and Pattern Recognition}}. \bibinfo{pages}{5778--5788}.
\newblock


\bibitem[Zhu et~al\mbox{.}(2021)]%
        {zhu_one_2021}
\bibfield{author}{\bibinfo{person}{Yuhao Zhu}, \bibinfo{person}{Qi Li},
  \bibinfo{person}{Jian Wang}, \bibinfo{person}{Cheng-Zhong Xu}, {and}
  \bibinfo{person}{Zhenan Sun}.} \bibinfo{year}{2021}\natexlab{}.
\newblock \showarticletitle{One shot face swapping on megapixels}. In
  \bibinfo{booktitle}{\emph{Proceedings of the IEEE/CVF Conference on Computer
  Vision and Pattern Recognition}}. \bibinfo{pages}{4834--4844}.
\newblock


\bibitem[Zi et~al\mbox{.}(2020)]%
        {zi_wilddeepfake_2020}
\bibfield{author}{\bibinfo{person}{Bojia Zi}, \bibinfo{person}{Minghao Chang},
  \bibinfo{person}{Jingjing Chen}, \bibinfo{person}{Xingjun Ma}, {and}
  \bibinfo{person}{Yu~Gang Jiang}.} \bibinfo{year}{2020}\natexlab{}.
\newblock \showarticletitle{{WildDeepfake}: {A} {Challenging} {Real}-{World}
  {Dataset} for {Deepfake} {Detection}}.
\newblock \bibinfo{journal}{\emph{MM 2020 - Proceedings of the 28th ACM
  International Conference on Multimedia}} (\bibinfo{year}{2020}),
  \bibinfo{pages}{2382--2390}.
\newblock
\urldef\tempurl%
\url{https://doi.org/10.1145/3394171.3413769}
\showDOI{\tempurl}
\newblock
\shownote{ISBN: 9781450379885}.


\end{thebibliography}

\end{document}